\documentclass[10pt,journal,compsoc]{IEEEtran}
\usepackage[switch]{lineno}

\ifCLASSOPTIONcompsoc
  \usepackage[nocompress]{cite}
\else
  \usepackage{cite}
\fi
\usepackage{hyperref} 
\hypersetup{
    colorlinks=true,
    linkcolor=red,
}
\usepackage{amssymb}
\usepackage{bbding}

\ifCLASSINFOpdf
   \usepackage[pdftex]{graphicx}
\else
\fi
\usepackage{amsmath,amsfonts}
\usepackage{array}
\usepackage[caption=false,font=normalsize,labelfont=sf,textfont=sf]{subfig}
\usepackage{textcomp}
\usepackage{stfloats}
\usepackage{url}
\usepackage{verbatim}
\usepackage{graphicx}
\usepackage{cite}
\usepackage{xcolor}
\usepackage{booktabs}
\usepackage{url}
\usepackage{amssymb}
\usepackage{multirow} 
\usepackage{fontawesome} 
\usepackage{algpseudocode}

\begin{document}

\title{Rethinking the Evaluation of Visible and Infrared Image Fusion}


\author{Dayan~Guan,
    Yixuan~Wu,
    Tianzhu~Liu, Alex C. Kot,~\IEEEmembership{Life Fellow,~IEEE}, \\
    and Yanfeng~Gu,~\IEEEmembership{Senior Member,~IEEE}
\IEEEcompsocitemizethanks{
\IEEEcompsocthanksitem 
Dayan~Guan, Tianzhu~Liu and Yanfeng~Gu are with the School of Electronics and Information Engineering, Harbin Institute of Technology, China.
\IEEEcompsocthanksitem 
Yixuan~Wu is with the School of Computer and Communication Engineering, Northeastern University, Qinhuangdao, China.
\IEEEcompsocthanksitem 
Alex C. Kot is with the School of Electrical and Electronic Engineering, Nanyang Technological University
\IEEEcompsocthanksitem 
Corresponding authors: Tianzhu~Liu (tzliu@hit.edu.cn) and Yanfeng~Gu (guyf@hit.edu.cn).
}
}


\markboth{Journal of \LaTeX\ Class Files,
March~2023}%
{Shell \MakeLowercase{\textit{et al.}}: Bare Demo of IEEEtran.cls for Computer Society Journals}



\IEEEtitleabstractindextext{%
\begin{abstract}
Visible and Infrared Image Fusion (VIF) has garnered significant interest across a wide range of high-level vision tasks, such as object detection and semantic segmentation. However, the evaluation of VIF methods remains challenging due to the absence of ground truth. This paper proposes a Segmentation-oriented Evaluation Approach (SEA) to assess VIF methods by incorporating the semantic segmentation task and leveraging segmentation labels available in latest VIF datasets. Specifically, SEA utilizes universal segmentation models, capable of handling diverse images and classes, to predict segmentation outputs from fused images and compare these outputs with segmentation labels. Our evaluation of recent VIF methods using SEA reveals that their performance is comparable or even inferior to using visible images only, despite nearly half of the infrared images demonstrating better performance than visible images. Further analysis indicates that the two metrics most correlated to our SEA are the gradient-based fusion metric $Q_{\text{ABF}}$ and the visual information fidelity metric $Q_{\text{VIFF}}$ in conventional VIF evaluation metrics, which can serve as proxies when segmentation labels are unavailable. We hope that our evaluation will guide the development of novel and practical VIF methods. The code has been released in \url{https://github.com/Yixuan-2002/SEA/}.
\end{abstract}

}

\maketitle

\section{Introduction}
\IEEEPARstart{I}{mages} captured by a single modal sensor often fail to provide a comprehensive and accurate depiction of the imaging scene due to inherent theoretical and technical limitations~\cite{liu2011objective,ma2019infrared,zhang2023visible,zhang2021image}. Infrared sensors, which detect thermal radiation emitted by objects, excel in highlighting prominent targets but lack color and texture information. Conversely, visible sensors capture reflective light information, producing images rich in color and texture details but are highly sensitive to environmental factors such as illumination and occlusion. These complementary characteristics underscore the potential of fusing infrared and visible images to create composite images that highlight prominent targets and preserves intricate details. Therefore, Visible and Infrared Image Fusion (VIF) has become increasinly prevalent as a pre-processing step in various high-level vision tasks, including object detection~\cite{hwang2015multispectral,guan2019fusion,zhao2023metafusion}, object tracking~\cite{zhang2020object,li2021lasher,zhang2021learning}, person re-identification~\cite{wang2019learning,yu2023modality,ye2023channel}, and semantic segmentation~\cite{zhou2021gmnet}. 

\begin{figure}[t]
    \centering
        \includegraphics[page=1, width=\linewidth]{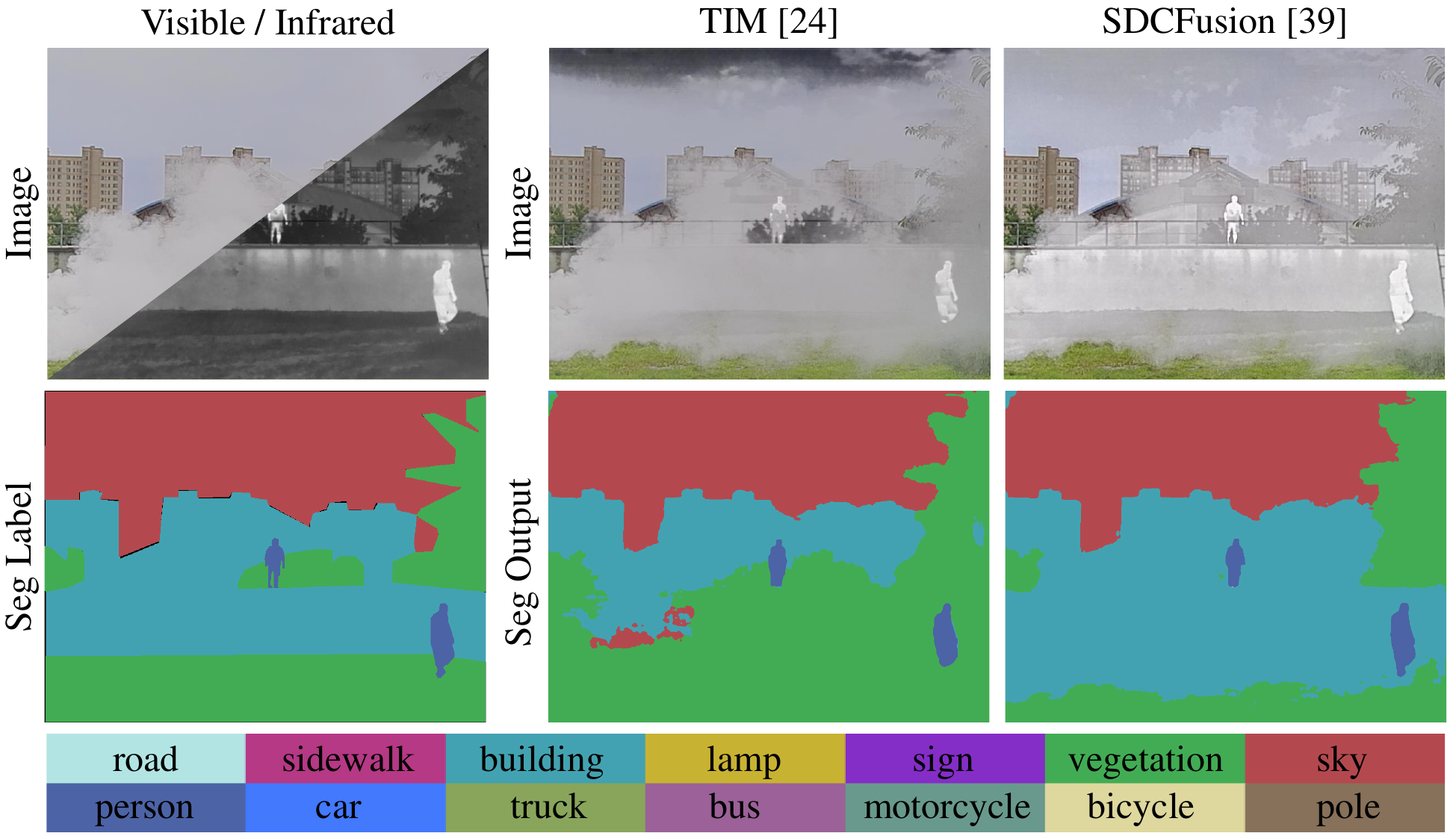}
        \vspace{-10pt}
        \caption{
        Evaluating the quality of fused images in VIF poses a significant challenge due to the lack of ground truth. To address this challenge, this paper proposes a novel segmentation-oriented evaluation approach that leverages a semantic segmentation task for assessing the quality of fused images. The underlying reason is that better segmentation performance indicates better fusion quality due to the intrinsic consistency between visual and semantic information~\cite{zhang2024mrfs}. 
        To illustrate, the first row shows the fused image generated from visible and infrared images using latest VIF methods TIM~\cite{liu2024task} and SDCFusion~\cite{liu2024semantic}, while the second row presents the corresponding segmentation label and outputs (from TIM and SDCFusion) predicted by the state-of-the-art universal segmentation model X-Decoder~\cite{zou2023generalized}, with the last row showing the color palette for different classes. 
        }
    \label{fig_motivation}
\end{figure}

Over the past years, numerous VIF techniques have been developed, evolving from traditional methods~\cite{kong2007multiscale,liu2015general,ma2016infrared,zhou2016perceptual,li2017pixel} to advanced deep learning-based approaches~\cite{xu2020u2fusion,li2023lrrnet,yi2024text,liu2024semantic,zhang2024mrfs}. While the latest deep learning-based methods have demonstrated the ability to produce high-quality fused images, several critical challenges persist within the image fusion community. Foremost among these is the difficulty in evaluating VIF methods due to the unavailability of reference fused images, commonly referred to as the ground truth, in real-world scenarios. 
To address this issue, recent methods have leveraged pixel-level segmentation labels available in many existing VIF datasets~\cite{ha2017mfnet,shivakumar2020pst900,ji2023semanticrt,liu2023multi,ji2023multispectral}. By using these labels, researchers have either trained additional segmentation models~\cite{tang2022piafusion,li2023deep,zhao2023cddfuse,zhang2024dispel,zhao2024equivariant} or developed unified models that perform both image fusion and segmentation~\cite{tang2022image,tang2022superfusion,tang2023rethinking,liu2023paif,liu2024task,liu2024semantic,zhang2024mrfs}. These semantic models are then used to assess the quality of the fused images, with better segmentation performance indicating better fusion quality due to the intrinsic consistency between visual and semantic information~\cite{zhang2024mrfs}. However, training semantic models on specific VIF datasets is impractical for evaluating VIF methods because these methods should be generalizable across different datasets, whereas the trained semantic models are often only applicable to the dataset they were trained on. This limitation underscores the need for universally applicable evaluation approach for VIF methods.

To this end, this paper introduces a Segmentation-oriented Evaluation Approach (SEA) that assesses VIF methods by using universal segmentation models to facilitate robust segmentation on datasets with different classes.  Specifically, the SEA forwards the class names and fused images from VIF methods to the pre-trained segmentation models to predict segmentation outputs, and compares these outputs with the annotated segmentation labels. The recent advancements in universal segmentation models~\cite{zou2023generalized,zou2024segment,ren2024grounded} have demonstrated their capability to produce reasonable segmentation results across diverse datasets, encompassing various image types and different classes. During the evaluation with SEA, better segmentation performance indicates higher fusion quality, reflecting the intrinsic consistency between visual and semantic information. For example, as shown in Figure~\ref{fig_motivation}, the state-of-the-art universal segmentation model X-Decoder~\cite{zou2023generalized} excels in generating satisfactory segmentation results for high-quality images but struggles with images of low visual quality, such as those occluded by smoke in the VIF methods TIM~\cite{liu2024task} and SDCFusion~\cite{liu2024semantic}. With the proposed evaluation approach, future development of VIF methods can be directed not only towards the effective combination of information from source images but also towards mitigating the adverse effects of low visual quality in the source images, thereby better facilitating downstream high-level vision tasks.

Based on the proposed SEA, we evaluate 30 recent VIF methods using the state-of-the-art universal segmentation models over the FMB~\cite{liu2023multi} and MVSeg~\cite{ji2023multispectral} datasets. 
Experimental results reveal that these methods perform comparable or even worse than using visible images only, despite infrared images showing better performance than visible images in 40.2\% and 5.2\% of the FMB and MVSeg datasets, respectively.
These findings highlight the critical need for further development in VIF methods to achieve substantial performance gains. 
In addition, we adopt 15 conventional evaluation metrics to assess VIF methods, providing a more comprehensive analysis than the previous VIF survey papers~\cite{liu2011objective,ma2019infrared,zhang2023visible} in terms of approach (incorporating more recent methods), criteria (utilizing a larger number of evaluation metrics), and data (including the latest datasets). Furthermore, we utilize a statistical correlation measure to assess the consistency between our SEA and conventional evaluation metrics. This correlation analysis shows that the two metrics most correlated to our SEA are the gradient-based fusion metric $Q_{\text{ABF}}$ and the visual information fidelity metric $Q_{\text{VIFF}}$ among conventional VIF evaluation metrics. Given the superiority of the proposed SEA, $Q_{\text{ABF}}$ and $Q_{\text{VIFF}}$ should be considered when the segmentation labels are unavailable. We hope that our evaluation can provide valuable insights into the development of novel and practical VIF methods, guiding future research to address current limitations and achieve significant performance improvements.

In summary, the contributions of this work are three-fold:
\begin{itemize}
\item It proposes a novel Segmentation-oriented Evaluation Approach (SEA) for assessing Visible and Infrared Image Fusion (VIF) methods by introducing a universal segmentation task. 
This approach addresses the challenge of ground-truth absence in VIF evaluation and is universally applicable across diverse VIF datasets, accommodating segmentation labels from different classes.
\item It performs a comparative study to evaluate the effectiveness of 30 recent VIF methods using the proposed SEA and 15 conventional evaluation metrics on the latest VIF datasets. This experimental study is more comprehensive than prior research in terms of involving more recent methods, evaluation metrics and latest datasets.
\item It conducts a correlation analysis by measuring the performance consistency between the proposed SEA and conventional evaluation metrics. This analysis indicates that conventional evaluation metrics with high correlation to SEA should be applied when semantic labels are inaccessible. 
\end{itemize}

The remainder of this paper is structured as follows: 
Section~\ref{Segmentation-oriented Evaluation Approach} presents the proposed evaluation approach in detail, explaining its methodology and implementation.
Section~\ref{Evaluation Datasets} describes the datasets used for evaluation, including their characteristics and relevance to this work.
Section~\ref{Evaluated Methods} outlines the methods evaluated in this paper, discussing the application of the SEA.
Section~\ref{Comparative Study} provides a comprehensive comparative study of recent open-source VIF methods using the proposed SEA.
Section~\ref{Correlation Analysis} explores the correlation analysis, examining the consistency between the SEA and conventional evaluation metrics.
Finally, Section~\ref{Conclusion} concludes the paper, summarizing the findings and suggesting directions for future research.

\section{Proposed Evaluation Approach}
\label{Segmentation-oriented Evaluation Approach}

Recent advancements in VIF methodologies~\cite{tang2022image,tang2022superfusion,tang2023rethinking,liu2023paif,liu2024task,liu2024semantic,zhang2024mrfs}  have successfully integrated semantic segmentation tasks to enhance the visual quality of fused images by exploiting the intrinsic consistency between visual and semantic information. Drawing inspiration from these studies, this paper introduces a Segmentation-oriented Evaluation Approach (SEA) that utilizes semantic segmentation to evaluate the visual quality of VIF-fused images.  The chosen segmentation model must be versatile, capable of handling diverse images and classes, to ensure compatibility with datasets comprising various image types/modalities and classes. To meet this criterion, we select three of the latest and most popular universal segmentation models: X-Decoder~\cite{zou2023generalized}, SEEM~\cite{zou2024segment}, and G-SAM~\cite{ren2024grounded}. The subsequent parts of this section will elaborate on why these universal segmentation models are capable of managing diverse images and classes.

\begin{figure}[!t]
    \centering
        \includegraphics[width=\linewidth]{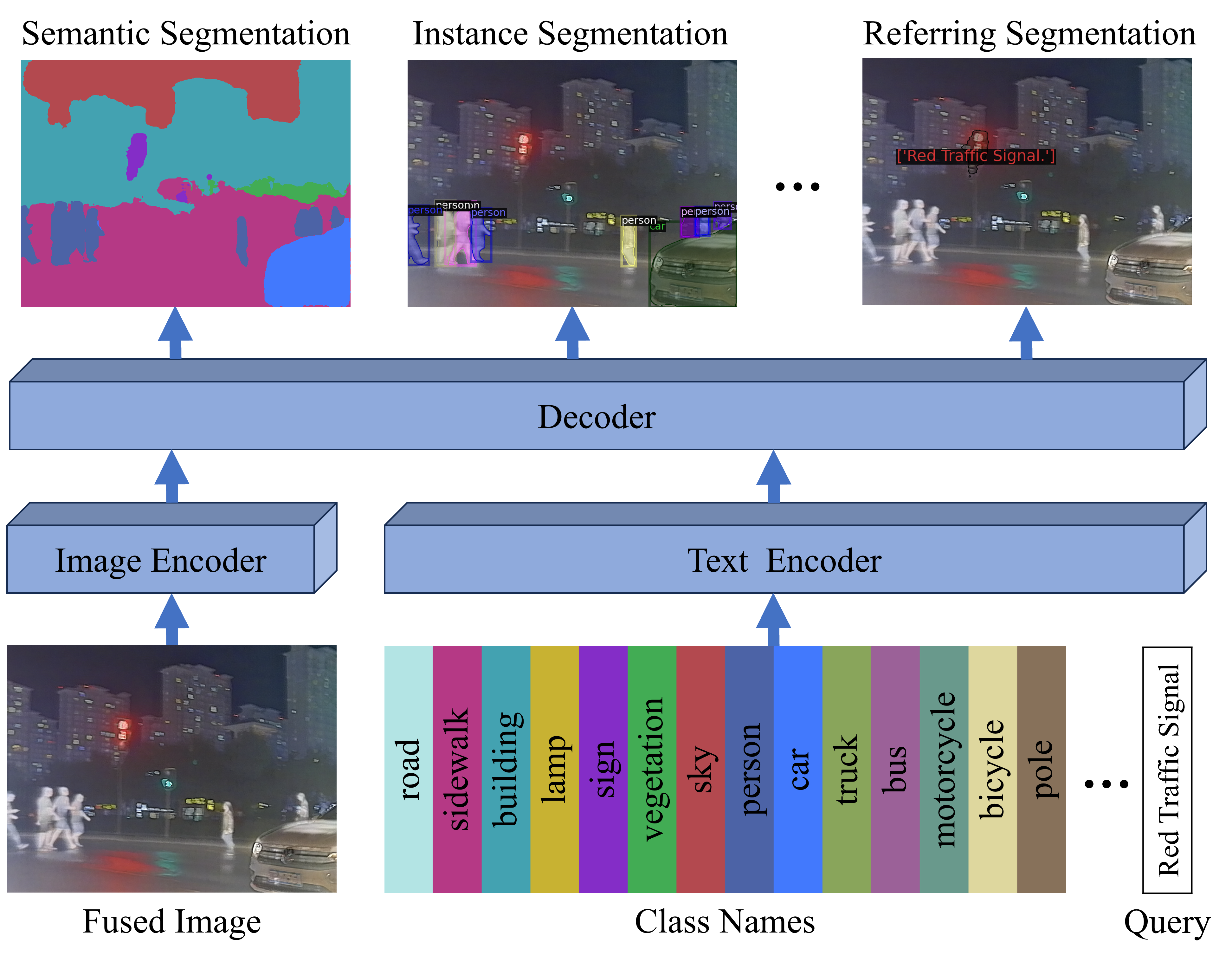}
        \vspace{-10pt}
        \caption{
        Overview of current universal segmentation models featuring an image/text encoder-decoder architecture. The encoders are designed to process diverse image inputs (across various styles and modalities) and text inputs (including different class names or queries). The decoder is capable of performing multiple high-level vision tasks, such as semantic segmentation, instance segmentation, referring segmentation, and etc. Our proposed SEA leverages the semantic segmentation task capabilities of current universal segmentation models to evaluate the quality of VIF fused images. 
        }
    \label{fig_universal_segmentation}
\end{figure}

\begin{figure*}[!h]
    \centering
        \includegraphics[width=\linewidth]{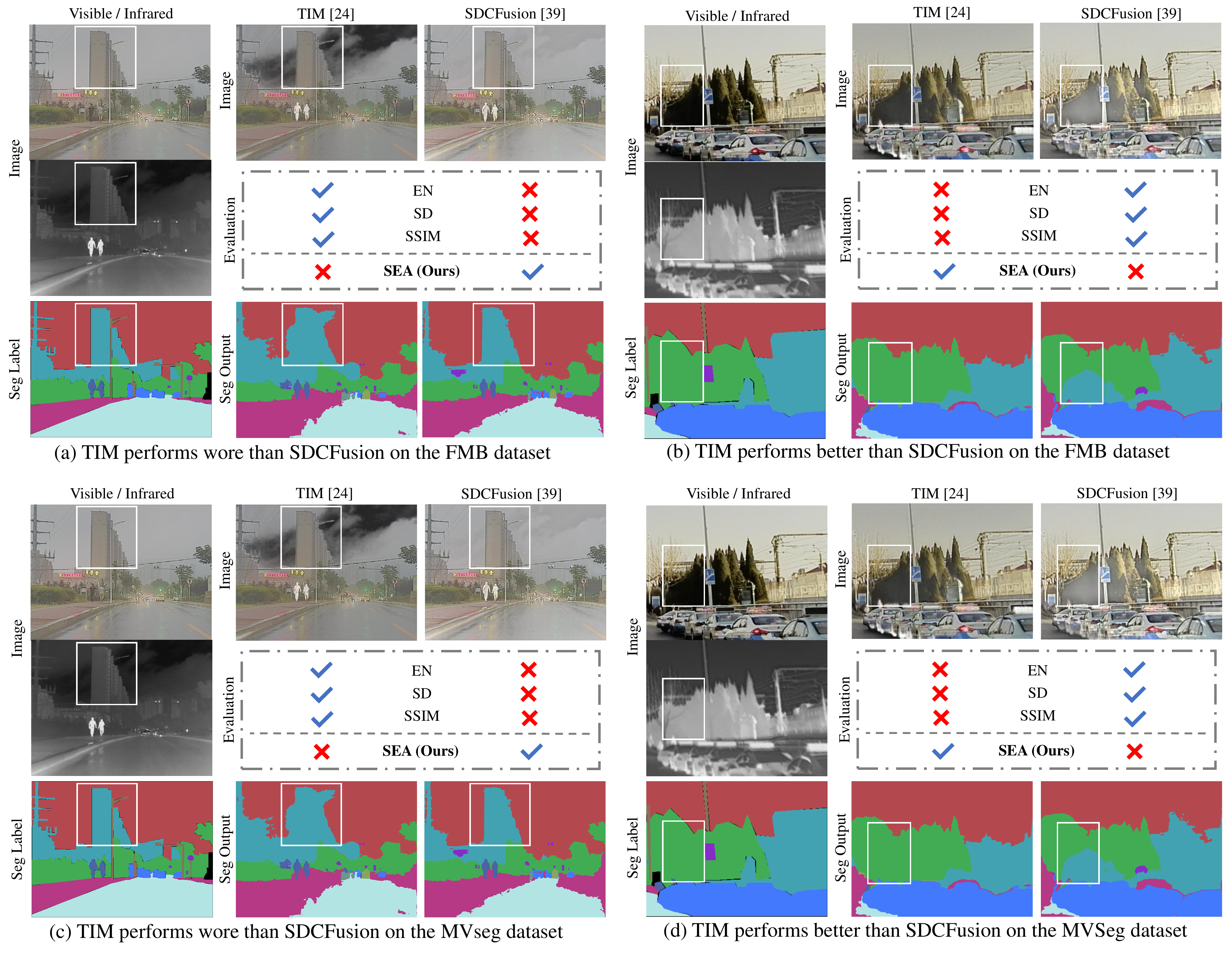}
        \vspace{-20pt}
        \caption{
        Image quality assessment of latest VIF methods (TIM~\cite{liu2024task} and SDCFusion~\cite{liu2024semantic}) using our proposed SEA alongside 3 widely used evaluation metrics including Entropy (EN), Standard Deviation (SD) and SSIM. Our SEA demonstrates superior performance on both the FMB and MVSeg datasets. Note that \textcolor{blue}{\faCheck} indicates better performance, while \textcolor{red}{\faTimes} indicates worse performance. 
        }
    \label{fig_sub_assess}
\end{figure*}

Current universal segmentation models are predominantly developed upon the foundational architecture of large vision-language models. Pre-trained large vision-language models, such as CLIP~\cite{radford2021learning}, have demonstrated significant promise in generating representations that can be effectively transferred to downstream classification tasks. Distinct from traditional representation learning, which largely depends on discretized labels, vision-language pre-training aligns images and texts within a unified feature space and this alignment facilitates zero-shot transfer to downstream tasks via text prompting, where classification weights are generated from natural language descriptions of the target classes.  Furthermore, by leveraging a training dataset with hundreds of millions of image-text pairs, the CLIP model demonstrates robust capabilities in recognizing and adapting to a wide array of image styles, including sketches that lack color and detail information. Consequently, the CLIP model excels in managing diverse image types and classes within the image classification task.

Universal segmentation models~\cite{zou2023generalized, zou2024segment,ren2024grounded,dong2024generative} extends the pre-trained large vision-language models~\cite{radford2021learning,karmanov2024efficient} from the image classification task to the semantic segmentation task. These models typically employ a vision-language encoder-decoder architecture, as illustrated in Figure~\ref{fig_universal_segmentation}. In this architecture, the vision and language encoders are  derived from pre-trained large vision-language models like CLIP, and the decoder is specifically designed to handle a variety of segmentation tasks, including semantic segmentation, instance segmentation, and referring segmentation. This approach offers three key advantages for managing diverse images and classes across different VIF datasets: First, the vision encoder's ability to handle various image modalities, such as infrared imagery, is bolstered by CLIP's proven effectiveness with sketch images, which similarly lack color and detail; Second, the language encoder can process different class names within various VIF segmentation datasets; Third, the decoder, trained in a multi-task learning manner~\cite{caruana1997multitask,Li2023AnEF,Gu2022AnIS,dong2024upetu}, enhances generalization by utilizing domain-specific information embedded in the training signals of related tasks. In evaluating fused images generated by VIF methods, we focus exclusively on the pixel-level semantic segmentation task. This choice leverages the intrinsic consistency between vision and semantics to assess visual quality at the pixel level.


To formally present the procedure of SEA in evaluating the performance of image fusion methods through a universal segmentation model $G$, we will now detail the methodological framework. For the sake of simplicity, this procedure will focus exclusively on the semantic segmentation outputs produced by $G$. Consider an image fusion model $F_{A}$ generated by a VIF method A. Given $N$ pairs of visible and infrared images $\{x^{V}_{i}, x^{I}_{i}\}_{i=1}^{N}$ from a VIF dataset, the image fusion model $F_{A}$ produces a fused image $x^{A}_{i} = F_{A}(x^{V}_{i}, x^{I}_{i})$ for each corresponding pair of visible and infrared images. Next, using the segmentation labels $\{\hat{y}_{i}\}_{i=1}^{N}$ and associated class names $c$ provided within the VIF dataset, SEA employs the universal segmentation model $G$ to predict segmentation outputs $y^{A}_{i} = G(x^{A}_{i}, c)$. As depicted in Figure~\ref{fig_universal_segmentation}, the universal segmentation model $G$ is composed of three core components: an image encoder $E_{I}$, a text encoder $E_{T}$, and a decoder $D$. Consequently, the segmentation output for each fused image can be expressed as: $y^{A}_{i}=G(x^{A}_{i},c)=D(E_{I}(x^{A}_{i}),E_{T}(c))$. To evaluate the performance of the VIF method A on the dataset, SEA calculates the mean Intersection over Union (mIoU) score, which is commonly used in the evaluation of segmentation methods. The performance score $s^{A}$ for method A is then determined by: $s^{A}=S(\{y^{A}_{i},\hat{y}_{i}\}_{i}^{N})$, where $S$ represents the mIoU computation function that compares the predicted segmentation outputs with the ground truth labels. 

Furthermore, we analyse the advancements of our proposed SEA compared with conventional evaluation metrics. For qualitative analysis, we select the latest VIF methods TIM~\cite{liu2024task} (published in the journal of IEEE TPAMI in 2024) and SDCFusion~\cite{liu2024semantic} (published in the journal of Information Fusion in 2024). 
For conventional evaluation, we select three widely used metrics: Entropy (EN), Standard Deviation (SD), and Structural Similarity Index (SSIM). Figure~\ref{fig_sub_assess} showcases some representative results on the FMB and MVSeg datasets.

As illustrated in Figures~\ref{fig_sub_assess}(a)~and~\ref{fig_sub_assess}(c), TIM performs worse than SDCFusion because the sky regions in TIM are influenced by infrared imagery, resulting in an ``unreasonable" black color during the daytime. In this scenario, our SEA provides a correct quality assessment by evaluating the semantic content, whereas EN, SD, and SSIM offer an incorrect assessment by simply judging the amount of information in the fused images, even if such information is noise. A similar phenomenon is observed in Figures~\ref{fig_sub_assess}(b)~and~\ref{fig_sub_assess}(d), where TIM outperforms SDCFusion. The color information of tree regions in SDCFusion is affected by infrared imagery, leading to a lack of color information, making it difficult for humans and intelligent machines to recognize these regions.

Therefore, compared with conventional evaluation metrics, our proposed SEA can guide the further development of VIF methods. It not only promotes the effective combination of information from source images but also mitigates the adverse effects of low visual quality in the source images, thereby better facilitating downstream high-level vision tasks.

\renewcommand\arraystretch{1.5}
\begin{table*}[!t]
\caption{
Details of existing VIF Datasets. Note that SS, DV and UAV are the abbreviations of surveillance camera, driving vehicle and unmanned aerial vehicle, respectively. The labeled ratio in both the FMB and MVSeg datasets is nearly 100\%.
}
\centering
\begin{tabular}{|l|c|c|c|c|c|c|c|c|}
\hline
Dataset &Segmenation &Labeled Ratio &Classes &Image Pairs &Resolution &Platform &Publication &Year \\
\hline
OSU~\cite{davis2007background} &\texttimes &- &- &285 &320$\times$240 &SS &CVIU &2007 \\ 
RGBT234~\cite{li2019rgb} &\texttimes &- &- &233.8K &640$\times$480 &SS &PATT RECOGN &2019 \\ 
LLVIP~\cite{jia2021llvip} &\texttimes &- &- &16,836 &1080$\times$720 &SS &ICCV &2021 \\ 
KAIST~\cite{hwang2015multispectral} &\texttimes &- &- &95K &640$\times$480 &DV &CVPR &2015 \\ 
Multispectral~\cite{takumi2017multispectral} &\texttimes &- &- &2,999 &768$\times$576 &DV &ACM MM &2017 \\ 
Roadscene~\cite{xu2020u2fusion} &\texttimes &- &- &221 &768$\times$576 &DV &IEEE TPAMI &2020 \\ 
M$^{3}$FD~\cite{liu2022target} &\texttimes &- &- &4,200 &1024$\times$768 &DV, SS &CVPR &2022 \\ 
VTUAV~\cite{zhang2022visible} &\texttimes &- &- &1.7M &1920$\times$1080 &UAV &CVPR &2022 \\ 
DroneVehicle~\cite{sun2022drone} &\texttimes &- &- &28,439 &640$\times$512 &UAV &IEEE TCSVT &2022 \\ 
RGBTDrone~\cite{zhang2023drone} &\texttimes &- &- &6,125 &640$\times$512 &UAV &ISPRS JPRS &2023 \\ 
\hline
MFNet~\cite{ha2017mfnet} &${\surd}$ &8.7\% &8 &1,569 &640$\times$480 &DV &IROS &2017 \\ 
PST900~\cite{shivakumar2020pst900} &${\surd}$ &3.0\% &5 &894 &1280$\times$720 &SS &ICRA &2020 \\ 
SemanticRT~\cite{ji2023semanticrt} &${\surd}$ & 21.5\% &13 &11,371 &Various &SS &ACM MM &2023 \\ \hline
FMB~\cite{liu2023multi} &${\surd}$ &98.2\% &14 &1,500 &800$\times$600 &DV, SS &CVPR &2023 \\ 
MVSeg~\cite{ji2023multispectral} &${\surd}$ &99.0\% &26 &3,545 &Various &DV, SS &CVPR &2023 \\ 
\hline
\end{tabular}
\label{existing_VIF_datasets}
\end{table*}

\section{Evaluation Datasets}
\label{Evaluation Datasets}

Recent advancements in VIF have been significantly propelled by the availability of public datasets, which provide standardized benchmarks. These benchmarks enable researchers to compare the performance of various fusion methods fairly and consistently. In this section, we describe the datasets used for evaluation, their characteristics, and their relevance to this work.

Semantic segmentation plays a pivotal role in the proposed segmentation-oriented evaluation approach. By leveraging segmentation labels, our evaluation approach can directly assess the quality of fused images based on their alignment with semantic information. The labeled ratio indicates the proportion of the dataset that includes annotated segmentation labels, which is critical for evaluating the effectiveness of VIF methods. High labeled ratios in the selected datasets ensure a reliable and extensive assessment of the fusion methods. The details of existing VIF datasets are shown in the Table~\ref{existing_VIF_datasets}.

In this paper, we selected FMB~\cite{liu2023multi} and MVSeg~\cite{ji2023multispectral} datasets due to their segmentation annotations with high labeled ratios, ensuring a thorough evaluation of fusion quality through semantic segmentation.

\vspace{2pt}
FMB~\cite{liu2023multi} stands out with a 98.2\% labeled ratio and includes 14 classes. It comprises 1,500 image pairs (280 for testing) captured using a smart multi-wave binocular imaging system with a resolution of 800×600. This dataset is a robust platform for testing fusion methods in both driving and surveillance contexts.

\vspace{2pt}
MVSeg~\cite{ji2023multispectral} offers a 99.0\% labeled ratio, encompassing 25 classes and 3,545 image pairs of various resolutions, with 926 pairs designated for testing. These image pairs are sourced from multiple existing datasets, including OSU, INO, RGBT234, and KAIST. The diverse domains covered by MVSeg make it ideal for evaluating the generalizability of VIF methods.

In summary, the selected datasets offer high labeled ratios and diverse class annotations, making them ideal for a comprehensive evaluation of VIF methods. These datasets facilitate a robust comparison of fusion techniques and their ability to maintain semantic integrity across different scenarios and image types. 
Compared to the latest VIF survey paper~\cite{zhang2023visible} published in the journal of IEEE TPAMI in 2023, which evaluates using the VIFB dataset containing only 21 image pairs, our study extends the evaluation to the FMB and MVSeg datasets with 280 and 926 image pairs, respectively. This significantly larger number of testing samples ensures more comprehensive and robust evaluations of the VIF methods.

\renewcommand\arraystretch{1.5}
\begin{table*}[!t]
\caption{Details of recent open-source VIF methods. Note that `Unified' means the method is a unified framework of image fusion and segmentation. The category of Fusion model is summarized for each VIF method.}
\centering
\resizebox{\textwidth}{!}{%
\begin{tabular}{|l|c|c|c|c|c|c|}
\hline
Method &Unified &Fusion Model &Segmentation Model &Link of the Source Code &Publication &Year \\
\hline
DenseFuse~\cite{li2018densefuse} &No &Autoencoder &- &\url{https://github.com/hli1221/imagefusion_densefuse} &IEEE TIP &2018 \\
FusionGAN~\cite{ma2019fusiongan} &No &GAN &- &\url{https://github.com/jiayi-ma/FusionGAN} &INF FUS &2019 \\
U2Fusion~\cite{xu2020u2fusion} &No &CNN &- &\url{https://github.com/hanna-xu/U2Fusion} &IEEE TPAMI &2020 \\
DDcGAN~\cite{ma2020ddcgan} &No &GAN &- &\url{https://github.com/hanna-xu/DDcGAN} &IEEE TIP &2020 \\
SDNet~\cite{zhang2021sdnet} &No &CNN &- &\url{https://github.com/HaoZhang1018/SDNet} &IJCV &2021 \\
RFNNest~\cite{li2021rfn} &No &Autoencoder &- &\url{https://github.com/hli1221/imagefusion-rfn-nest} &INF FUS &2021 \\
SwinFusion~\cite{ma2022swinfusion} &No &Transformer &- &\url{https://github.com/Linfeng-Tang/SwinFusion} &IEEE JAS &2022 \\
PIAFusion~\cite{tang2022piafusion} &No &CNN &BAD~\cite{peng2021bilateral} &\url{https://github.com/Linfeng-Tang/PIAFusion} &INF FUS &2022 \\
TarDAL~\cite{liu2022target} &No &GAN &- &\url{https://github.com/JinyuanLiu-CV/TarDAL} &CVPR &2022 \\
LRRNet~\cite{li2023lrrnet} &No  &CNN & - &\url{https://github.com/hli1221/imagefusion-LRRNet} &IEEE TPAMI &2023 \\
DifFusion~\cite{yue2023dif} &No &Diffusion Model &- &\url{https://github.com/GeoVectorMatrix/Dif-Fusion} &IEEE TIP &2023 \\
TGFuse~\cite{rao2023tgfuse} &No &Transformer &- &\url{https://github.com/dongyuya/TGFuse} &IEEE TIP &2023 \\
DIVFusion~\cite{tang2023divfusion} &No &CNN &- &\url{https://github.com/Xinyu-Xiang/DIVFusion} &INF FUS &2023 \\
DLF~\cite{li2023deep} &No &Transformer &Bisenet-v2~\cite{yu2021bisenet} &\url{https://github.com/yuliu316316/IVF-WoReg} &IJCV &2023 \\
CDDFuse~\cite{zhao2023cddfuse} &No &Transformer-CNN &DeeplabV3+~\cite{chen2018encoder} &\url{https://github.com/Zhaozixiang1228/MMIF-CDDFuse} &CVPR &2023 \\
MetaFusion~\cite{zhao2023metafusion} &No &CNN &- &\url{https://github.com/wdzhao123/MetaFusion} &CVPR &2023 \\
DDFM~\cite{zhao2023ddfm} &No &Diffusion Model &- &\url{https://github.com/Zhaozixiang1228/MMIF-DDFM} &ICCV &2023 \\
SHIP~\cite{zheng2024probing} &No &CNN &- &\url{https://github.com/zheng980629/ship} &CVPR &2024 \\
TCMoA~\cite{zhu2024task} &No &Tranformer &- &\url{https://github.com/YangSun22/TC-MoA} &CVPR &2024 \\
TextIF~\cite{yi2024text} &No &Tranformer &- &\url{https://github.com/XunpengYi/Text-IF} &CVPR &2024 \\
DDBF~\cite{zhang2024dispel} &No &GAN &SegNeXt~\cite{guo2022segnext} &\url{https://github.com/HaoZhang1018/DDBF} &CVPR &2024 \\
EMMA~\cite{zhao2024equivariant} &No &Tranformer &DeeplabV3+~\cite{chen2018encoder}  &\url{https://github.com/Zhaozixiang1228/MMIF-EMMA} &CVPR &2024 \\
\hline
SeAFusion~\cite{tang2022image} &Yes &CNN &BAD~\cite{peng2021bilateral} &\url{https://github.com/Linfeng-Tang/SeAFusion} &INF FUS &2022 \\
SuperFusion~\cite{tang2022superfusion} &Yes &CNN &BAD~\cite{peng2021bilateral} &\url{https://github.com/Linfeng-Tang/SuperFusion} &IEEE JAS &2022 \\
PSFusion~\cite{tang2023rethinking} &Yes &CNN &SegNeXt~\cite{guo2022segnext} &\url{https://github.com/Linfeng-Tang/PSFusion} &INF FUS &2023 \\
SegMiF~\cite{liu2023multi} &Yes &CNN & SegFormer~\cite{xie2021segformer} &\url{https://github.com/JinyuanLiu-CV/SegMiF} &ICCV &2023 \\
PAIF~\cite{liu2023paif} &Yes &Transformer-CNN &SegFormer~\cite{xie2021segformer} &\url{https://github.com/LiuZhu-CV/PAIF} &ACM MM &2023 \\
TIM~\cite{liu2024task} &Yes &CNN &ABMDRNet~\cite{zhang2021abmdrnet} &\url{https://github.com/liuzhu-cv/timfusion} &IEEE TPAMI &2024 \\
SDCFusion~\cite{liu2024semantic} &Yes &CNN &UNet~\cite{ronneberger2015u} &\url{https://github.com/XiaoW-Liu/SDCFusion} &INF FUS &2024 \\
MRFS~\cite{zhang2024mrfs} &Yes &CNN &SegFormer~\cite{xie2021segformer} &\url{https://github.com/HaoZhang1018/MRFS} &CVPR &2024 \\
\hline
\end{tabular}
}
\label{VIF_methods}
\end{table*}

\begin{figure*}[h]
    \centering
        \includegraphics[width=\linewidth]{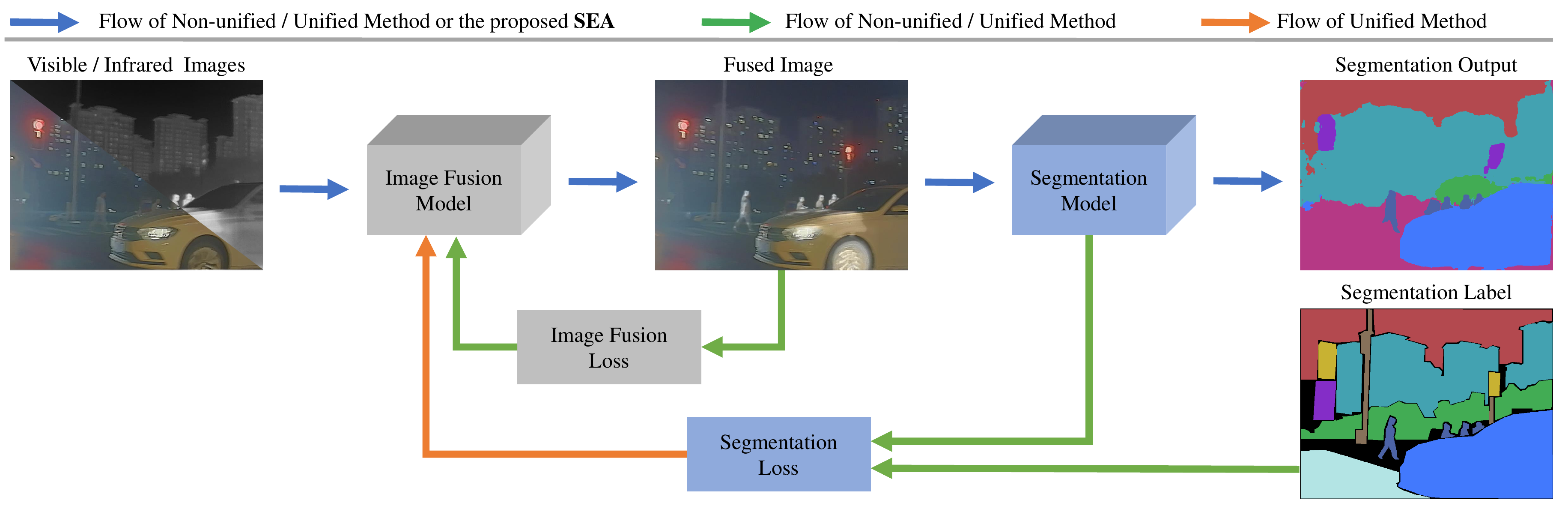}
        \vspace{-10pt}
        \caption{
        Comparisons of segmentation frameworks used in non-unified/unified VIF methods and our proposed SEA. Non-unified VIF methods involve training separate models for image fusion and segmentation, using the latter for subsequent evaluation. Conversely, integrate the training of fusion and segmentation models, employing a segmentation loss to refine the fusion process. In a comparison, our proposed SEA eliminates the requirement for additional training of the segmentation model.
        }
    \label{fig_framework}
\end{figure*}

\section{Evaluated Methods}
\label{Evaluated Methods}

Over recent years, numerous Visible and Infrared Image Fusion (VIF) techniques have evolved from traditional machine learning approaches to advanced deep learning methods. This paper focuses exclusively on the latest deep learning-based techniques that have demonstrated superior capabilities in producing high-quality fused images. 
Table~\ref{VIF_methods} provides a summary of the latest open-source VIF methods evaluated in this study. In contrast to the most recent VIF survey paper~\cite{zhang2023visible} published in the journal of IEEE TPAMI in 2023, which evaluated VIF methods proposed before September 2022, our paper includes a number of methods introduced in the past two years (2023-2024). Notably, these newly evaluated methods feature contributions from top-tier journals (2 in IEEE TPAMI, 2 in IEEE TIP) and leading conferences (8 in CVPR, 2 in ICCV). The recent VIF methods are categorized based on their fusion models into the following types: autoencoder-based, GAN-based, diffusion-model-based, CNN-based, and transformer-based approaches.

Autoencoder-based methods utilize auto-encoders~\cite{kramer1991nonlinear,hinton2011transforming,rezende2014stochastic} for feature extraction and reconstruction, employing specific fusion strategies for feature fusion. DenseFuse~\cite{li2018densefuse} is a pioneering method in this category, using an autoencoder for image reconstruction and applying various fusion rules for feature fusion. RFNNest~\cite{li2021rfn} improves upon this by integrating a residual fusion network that learns fusion rules through training with visible and infrared image pairs.

GAN-based methods incorporate generative adversarial mechanisms~\cite{goodfellow2014generative,mirza2014conditional,miyato2018spectral} into the VIF domain. FusionGAN~\cite{ma2019fusiongan} is the first method to use a generator for producing fused images with enhanced targets and a discriminator to ensure these images contain more textual details from visible images. DDcGAN~\cite{ma2020ddcgan} extends this approach with dual discriminators to preserve features from both source images, while TarDAL~\cite{liu2022target} and DDBF~\cite{zhang2024dispel} introduce novel techniques like joint training strategies and conditional generative adversarial networks to further refine fusion quality.

Diffusion-model-based methods capitalize on the capabilities of diffusion models~\cite{sohl2015deep,ho2020denoising,ho2020denoising} to generate images with higher quality than previous generative adversarial networks. DifFusion~\cite{yue2023dif} models the reconstruction of four-channel stacked images through the operation of diffusion, and DDFM~\cite{zhao2023ddfm} designs a post-sampling strategy built on a diffusion model for VIF, achieving fused image sampling using the well-structured DDPM~\cite{ho2020denoising} with no need of additional training.

CNN-based methods are known for their ability to perform feature extraction, fusion, and reconstruction, achieving superior results through innovative designs of network architectures. U2Fusion~\cite{xu2020u2fusion} designs an unsupervised CNN for VIF, enforcing the likeness between the fused images and visible/infrared images. SDNet~\cite{zhang2021sdnet} develops a squeeze-and-decomposition network to simultaneously conduct fusion and decomposition stages. PIAFusion~\cite{tang2022piafusion} proposes an illumination-guided model capable of fusing significant information from source images by identifying lighting variations. MetaFusion~\cite{zhao2023metafusion} introduces a meta-attribute embedding architecture closing the distance between image fusion and object detection. LRRNet~\cite{li2023lrrnet} proposes a neural network design approach for VIF, guided by network architecture search, optimizing the design process for superior fusion performance. SHIP~\cite{zheng2024probing} simulates complex interactions across different dimensions in the CNN, thereby significantly enhancing the collaboration between visible and infrared modalities.

Transformer-based methods exploit the transformer structure's capability to manage long-term connections within images. SwinFusion~\cite{ma2022swinfusion} designs an attention-guided cross-modality network for extensive combination of enhanced details and universal interaction. TGFuse~\cite{rao2023tgfuse} formulates a VIF algorithm by integrating transformer models and adversarial networks. CDDFuse~\cite{zhao2023cddfuse} introduces a two-stream Transformer-CNN architecture to extract and fuse both long-term and short-term features.

Apart from designing fusion models, latest VIF methods have either incorporated additional segmentation models in a non-unified manner or developed unified models that simultaneously perform image fusion and segmentation, as shown in Figure~\ref{fig_framework}. The non-unified methods train additional segmentation models to access their VIF performance, while the unified methods leverage pixel-level segmentation labels available in many existing VIF datasets to enhance their performance.

\vspace{2pt}
\textbf{Non-unified VIF Methods:} 
Several recent VIF methods, including PIAFusion~\cite{tang2022piafusion}, DLF~\cite{li2023deep}, CDDFuse~\cite{zhao2023cddfuse}, DDBF~\cite{zhang2024dispel} and EMMA~\cite{zhao2024equivariant}, have utilized additional segmentation models in a non-unified manner for performance evaluation. These VIF methods applied different segmentation models, such as BAD~\cite{peng2021bilateral}, Bisenet-v2~\cite{yu2021bisenet}, DeeplabV3+~\cite{chen2018encoder} and SegNeXt~\cite{guo2022segnext}, to train on the MFNet dataset.
These evaluation processes have three disadvantages: First, training identical segmentation models for all compared methods to ensure fair comparisons is a time-intensive process; Second, the MFNet dataset has a very low labeled ratio (only 8.7\%), which limits the ability to assess the majority of regions in the fused images; Third, the limited number of training samples in the MFNet dataset (only 784 pairs) increases the risk of over-fitting for the segmentation models.

\vspace{2pt}
\textbf{Unified VIF Methods:} 
Numerous recent VIF methods advocate for a unified framework that integrates image fusion and segmentation, aiming to overcome the limitations of previous approaches that neglected semantic information from high-level vision tasks. SeAFusion~\cite{tang2022image} introduces a semantic-aware efficient network by utilizing the real-time segmentation model BAD~\cite{peng2021bilateral} to provide high-level semantic features. 
Extending this approach, SuperFusion~\cite{tang2022superfusion} integrate image registration, fusion, and segmentation into a single unified architecture. 
PSFusion~\cite{tang2023rethinking} enhances VIF features by progressively injecting semantic features from the segmentation model SegNeXt~\cite{guo2022segnext}. 
SegMiF~\cite{liu2023multi} addresses the representation mismatch between a CNN-based fusion network and the segmentation model SegFormer~\cite{xie2021segformer} through a multilevel collaborative attention model. 
PAIF~\cite{liu2023paif} proposes a cognition-driven fusion network to enhance segmentation robustness in challenging environments. 
TIM~\cite{liu2024task} designs a regulated scheme to integrate features extracted from ABMDRNet~\cite{zhang2021abmdrnet}, guiding the unsupervised training procedure of VIF.
SDCFusion~\cite{liu2024semantic} designs a cross-domain interaction module to bolster the robustness of cross-modality coupled features extracted from a CNN-based fusion network and the segmentation model UNet~\cite{ronneberger2015u}. 
MRFS~\cite{zhang2024mrfs} constructs a mutual reinforcement between image fusion and segmentation, resulting in dual performance improvements in both tasks.
Evaluating VIF performance through integrated segmentation models in unified VIF methods presents three challenges: First, the training image pairs in these methods, similar to those in non-unified approaches, lack informativeness due to the limited size of current VIF datasets, increasing the risk of model over-fitting; Second, VIF models enhanced by segmentation models trained on these small-scale datasets often result in less generalizable fused images; Their, fair comparisons are hard to achieve since existing unified VIF methods utilize different segmentation models, and retraining these models within a unified framework is complicated and impractical.

This paper addresses the limitations of segmentation evaluation procedures in both non-unified and unified VIF methods through several key aspects:  
First, the datasets utilized in this study feature over 98\% labeled ratios, enabling comprehensive assessment of entire regions within fused images;
Second, the proposed SEA ensures fair comparisons by employing universal segmentation models that are generalizable across diverse image types and classes;
Finally, the evaluation process is highly efficient, as it does not require additional training of segmentation models.
By leveraging these advantages, this approach can promote the development of generalizable VIF models that not only exhibit high visual quality but also enhance downstream vision tasks.

\renewcommand\arraystretch{1.36}
\begin{table}[!t]
\centering
\caption{Quantitative comparisons of different VIF methods applied SEA on the FMB dataset using three universal segmentation models: SEEM, X-Decoder, and G-SAM. Results that exceed Visible by more than 1.0 mIoU are highlighted in \textbf{bold}.}
\begin{tabular}{|l|c|ccc|c|}
\hline
Method & Unified & SEEM & X-Decoder & G-SAM & Mean \\ \hline
Visible & No & 50.5 & 50.7 & 51.5 & 50.9  \\ 
Infrared & No & 43.5 & 42.3 & 41.4 & 42.4  \\ \hline
DenseFuse & No & 50.4 & 49.9 & 49.4 & 49.9  \\ 
FusionGAN & No & 44.3 & 37.7 & 42.2 & 41.4  \\ 
U2Fusion & No & 50.6 & 51.8 & 53.5 & \textbf{52.0}  \\ 
DDcGAN & No & 47.7 & 46.9 & 44.6 & 46.4  \\ 
SDNet & No & 47.8 & 47.6 & 45.1 & 46.8  \\ 
RPNNest & No & 48.5 & 47.4 & 47.7 & 47.9  \\ 
SwinFusion & No & 49.8 & 48.9 & 46.8 & 48.5  \\ 
PIAFusion & No & 51.8 & 52.6 & 50.9 & 51.8  \\ 
LRRNet & No & 50.0 & 49.3 & 48.7 & 49.3  \\ 
TarDAL & No & 49.7 & 48.8 & 48.6 & 49.0  \\ 
DifFusion & No & 50.1 & 51.7 & 48.1 & 50.0  \\ 
TGFuse & No & 50.7 & 48.7 & 47.5 & 49.0  \\ 
DIVFusion & No & 50.2 & 51.3 & 47.4 & 49.6  \\ 
DLF & No & 49.3 & 46.6 & 45.3 & 47.1  \\ 
CDDFuse & No & 52.7 & 53.0 & 50.9 & \textbf{52.2}  \\ 
MetaFusion & No & 49.4 & 51.6 & 51.5 & 50.8  \\ 
DDFM & No & 43.0 & 44.5 & 46.2 & 44.6  \\ 
SHIP & No & 52.1 & 51.0 & 49.1 & 50.7  \\ 
TCMoA & No & 51.9 & 49.7 & 47.9 & 49.8  \\ 
TextIF & No & 52.5 & 52.7 & 50.2 & 51.8  \\ 
DDBF & No & 48.2 & 51.2 & 44.3 & 47.9  \\ 
EMMA & No & 51.3 & 52.2 & 50.1 & 51.2  \\ \hline
SeAFusion & Yes & 50.9 & 51.4 & 49.6 & 50.6  \\ 
SuperFusion & Yes & 51.8 & 51.6 & 50.3 & 51.2  \\ 
PSFusion & Yes & 50.4 & 51.8 & 48.6 & 50.3  \\ 
SegMiF & Yes & 51.8 & 52.5 & 50.7 & 51.7  \\ 
PAIF & Yes & 50.9 & 52.1 & 51.1 & 51.4  \\ 
TIM & Yes & 50.4 & 51.2 & 48.6 & 50.1  \\ 
SDCFusion & Yes & 52.0 & 52.6 & 52.7 & \textbf{52.4}  \\ 
MRFS & Yes & 51.3 & 51.1 & 50.0 & 50.8  \\ \hline
\end{tabular}
\label{tab_fmb}
\end{table}

\renewcommand\arraystretch{1.36}
\begin{table}[!t]
\centering
\caption{Quantitative comparisons of different VIF methods applied SEA on the MVSeg dataset using three universal segmentation models: SEEM, X-Decoder, and G-SAM. Results that exceed Visible by more than 1.0 mIoU are highlighted in \textbf{bold}.}
\begin{tabular}{|l|c|ccc|c|}
\hline
Method & Unified & SEEM & X-Decoder & G-SAM & Mean \\ \hline
Visible & No & 18.5 & 20.0 & 24.7 & 21.1  \\ 
Infrared & No & 6.6 & 9.2 & 9.3 & 8.4  \\ \hline
DenseFuse & No & 17.6 & 19.0 & 23.7 & 20.1  \\ 
FusionGAN & No & 7.6 & 8.1 & 14.1 & 9.9  \\ 
U2Fusion & No & 18.0 & 19.4 & 24.4 & 20.6  \\ 
DDcGAN & No & 11.5 & 13.1 & 17.6 & 14.1  \\ 
SDNet & No & 14.3 & 15.4 & 19.0 & 16.2  \\ 
RPNNest & No & 15.4 & 17.4 & 23.2 & 18.7  \\ 
SwinFusion & No & 15.9 & 17.2 & 21.0 & 18.0  \\ 
PIAFusion & No & 17.5 & 18.6 & 23.5 & 19.9  \\ 
LRRNet & No & 16.2 & 16.6 & 23.2 & 18.7  \\ 
TarDAL & No & 14.2 & 14.5 & 20.4 & 16.4  \\ 
DifFusion & No & 16.9 & 17.3 & 22.5 & 18.9  \\ 
TGFuse & No & 16.3 & 18.0 & 22.7 & 19.0  \\ 
DIVFusion & No & 16.6 & 18.2 & 22.2 & 19.0  \\ 
DLF & No & 13.5 & 15.4 & 20.6 & 16.5  \\ 
CDDFuse & No & 17.2 & 18.2 & 23.3 & 19.6  \\ 
MetaFusion & No & 16.6 & 19.0 & 24.3 & 20.0  \\ 
DDFM & No & 14.7 & 16.4 & 21.4 & 17.5  \\ 
SHIP & No & 16.7 & 17.3 & 21.5 & 18.5  \\ 
TCMoA & No & 14.2 & 15.6 & 21.5 & 17.1  \\ 
TextIF & No & 18.2 & 19.6 & 24.7 & 20.8  \\ 
DDBF & No & 16.3 & 17.3 & 20.3 & 18.0  \\ 
EMMA & No & 16.8 & 18.2 & 23.7 & 19.6  \\ \hline
SeAFusion & Yes & 17.7 & 18.3 & 21.5 & 19.2  \\ 
SuperFusion & Yes & 16.6 & 17.6 & 22.3 & 18.8  \\ 
PSFusion & Yes & 17.2 & 19.0 & 23.7 & 20.0  \\ 
SegMiF & Yes & 17.3 & 18.3 & 22.5 & 19.4  \\ 
PAIF & Yes & 15.5 & 17.0 & 21.4 & 18.0  \\ 
TIM & Yes & 17.1 & 19.2 & 24.0 & 20.1  \\ 
SDCFusion & Yes & 18.0 & 19.2 & 23.4 & 20.2  \\ 
MRFS & Yes & 16.7 & 17.1 & 23.3 & 19.0  \\ \hline
\end{tabular}
\label{tab_mvseg}
\end{table}

\begin{figure}[t]
    \centering
        \includegraphics[width=\linewidth]{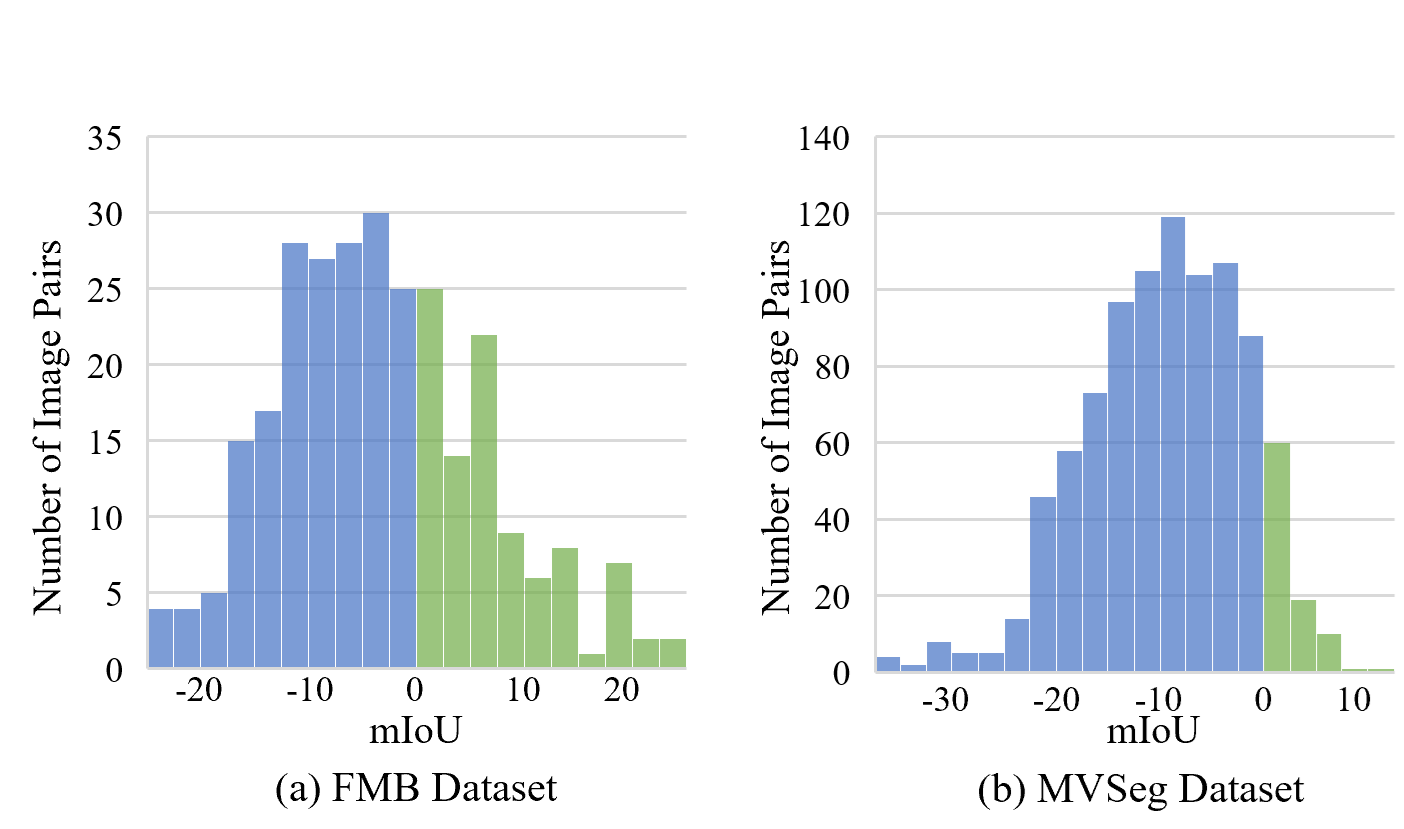}
        \vspace{-20pt}
        \caption{
        Performance differences between infrared and visible images. The mIoU is computed by subtracting the performance of each infrared image from that of its corresponding visible image across the FMB and MVSeg datasets. Green bars indicate cases where infrared images outperform visible images, while blue bars represent the opposite scenario.
        }
    \label{fig_30}
\end{figure}

\begin{figure}[t]
    \centering
        \includegraphics[width=\linewidth]{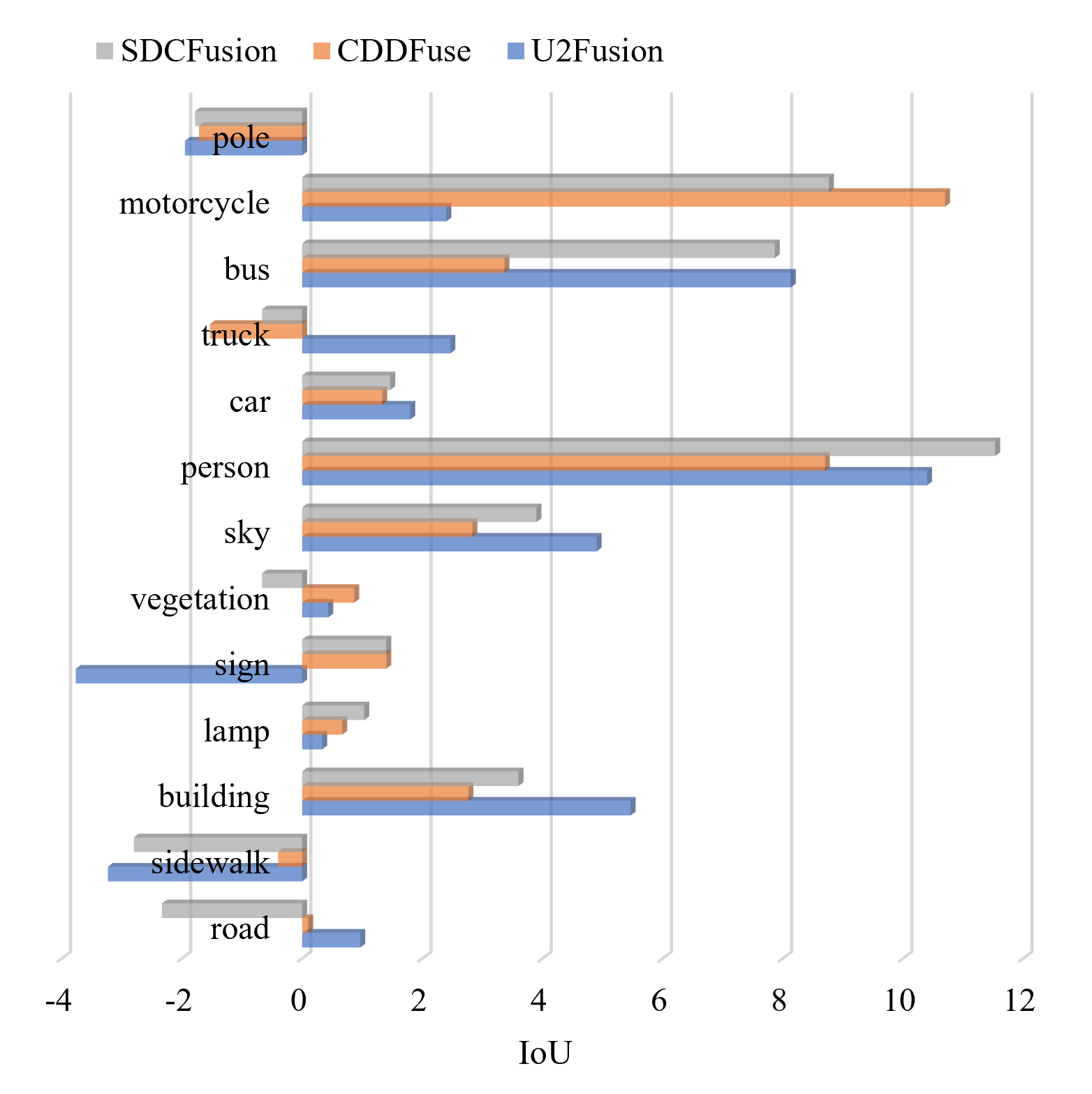}
        \vspace{-20pt}
        \caption{
        Comparative performance improvements of the VIF methods SDCFusion, CDDFusion, and U2Fusion, relative to using only the Visible images, evaluated for each class on the FMB dataset.
        }
    \label{fig_31}
\end{figure}

\begin{figure}[t]
    \centering
        \includegraphics[width=\linewidth]{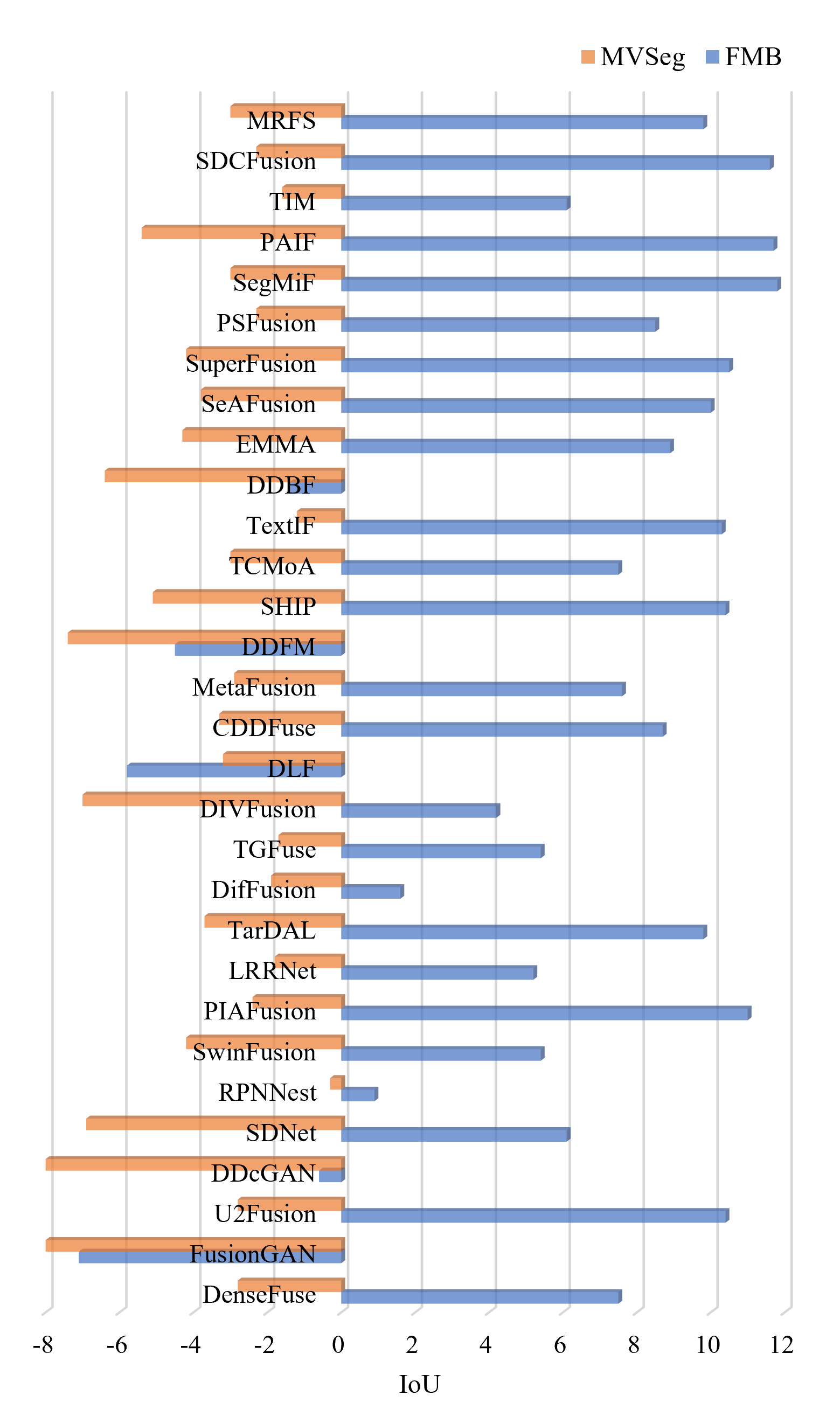}
        \vspace{-20pt}
        \caption{
        Comparative performance improvements of recent VIF methods, relative to using only the Visible images, evaluated for the person class on the FMB and MVSeg datasets. 
        }
    \label{fig_32}
\end{figure}

\section{Comparative Study}
\label{Comparative Study}

In this section, we apply our proposed SEA to evaluate 30 recent open-source VIF methods, as detailed in Table~\ref{VIF_methods}, utilizing three comprehensive segmentation models: SEEM, X-Decoder, and G-SAM. The quantitative results on the FMB dataset are presented in Table~\ref{tab_fmb}, revealing several key observations: 
First, only 3 methods (U2Fusion, CDDFuse, and SDCFusion) demonstrate a clear performance improvement (exceeding a 1.0 increase in mIoU) over the Visible in the Mean across all segmentation models; 
Second, unified models trained on VIF datasets with segmentation labels do not show superior segmentation results compared to non-unified models trained on VIF datasets without segmentation labels,  unified CNN-based models such as TIM and MRFS exhibit worse quantitative performance compared to non-unified CNN-based models like PIAFusion and MetaFusion; 
Third, generative VIF methods, particularly those based on GANs (FusionGAN, DDcGAN, TarDAL, DDBF) and diffusion models (DifFusion and DDFM), exhibit poor performance, with all methods performing noticeably worse than using the Visible images (by more than 1.0 in mIoU), and the GAN-based method FusionGAN even under-performing the Infrared; 
Finally, the latest VIF methods do not show advantages over older VIF methods, for instance, the latest methods such as DDBF and MRFS (published in 2024) show inferior quantitative performance compared to older methods like DenseFuse (published in 2019) and U2Fusion (published in 2020). 

Besides evaluating the proposed SEA on the FMB dataset, we extended our analysis to include the MVSeg dataset, which offers a larger set of testing samples and evaluation classes. The quantitative results of this evaluation are summarized in Table~\ref{tab_mvseg}. Notably, the trends observed in Table~\ref{tab_fmb} largely hold true for the MVSeg dataset as well. 
However, it is important to highlight that none of the methods evaluated showed a performance improvement over the Visible images. This lack of improvement underscores a significant limitation: existing VIF methods exhibit poor generalizability and struggle to effectively handle the diverse types of images sourced from multiple VIF datasets within the MVSeg dataset.

Furthermore, we explore whether the performance of recent VIF methods is impacted by the underperformance of infrared images. As shown in Figure~\ref{fig_30}, we compare the performance differences between infrared and visible images by calculating the mIoU difference between each infrared image and its corresponding visible image across the FMB and MVSeg datasets. The experimental results indicate that infrared images outperform visible images in 40.2\% of the FMB dataset and 5.2\% of the MVSeg dataset. These findings suggest that there is substantial potential to enhance recent VIF methods, as infrared images can provide more informative content than visible images in many scenarios.

Unlike conventional evaluation metrics that are class-agnostic, our proposed SEA is class-aware, enabling detailed analysis of semantic parts (regions of different classes) in the fused images. Initially, we evaluate various VIF methods on the FMB dataset, which comprises 14 classes. We identify three methods (U2Fusion, CDDFuse, and SDCFusion) that demonstrated a notable performance improvement, exceeding a 1.0 increase in mIoU compared to the Visible modality. As illustrated in Figure \ref{fig_31}, these methods achieve significant performance gains (over 8 in IoU) for the person class. This improvement is particularly relevant in VIF, as infrared images can provide clear signals for the person class when visibility is compromised in dark conditions.
To understand the consistency of these improvements, we further examined whether other VIF methods and datasets exhibit similar performance enhancements for the person class. As shown in Figure \ref{fig_32}, on the FMB dataset, four methods (FusionGAN, DLF, DDFM, DDBF) failed to achieve performance gains for the person class using additional infrared information. Interestingly, on the MVSeg dataset, none of the methods demonstrated similar performance improvements. 

Based on the observations above, we draw the following key insights:
First, most existing VIF methods fail to enhance visible imagery by integrating infrared information, even for the fused images with person class that exhibit distinct thermal signals.
Second, generative approaches, including GANs and diffusion models, prove unsuitable for the VIF task as they tend to compromise the semantic integrity necessary for downstream segmentation tasks.
Finally, existing VIF methods demonstrate poor generalization capabilities, limiting their effectiveness across diverse environmental conditions.
These findings underscore the urgent need for advancements in novel VIF methods to achieve meaningful improvements in performance and reliability.

\renewcommand\arraystretch{1.5}
\begin{table*}[!t]
\caption{Conventional evaluation metrics applied in recent VIF papers. The most widely used evaluation metrics are EN, SD, and SSIM. }
\centering
\resizebox{\textwidth}{!}{%
\begin{tabular}{|l|cccc|cccc|cccc|ccc|c|}
\hline
\multirow{2}{*}{Method} & \multicolumn{4}{c|}{Information theory-based} & \multicolumn{4}{c|}{Information feature-based} & \multicolumn{4}{c|}{Structural similarity-based} & \multicolumn{3}{c|}{Human perception-inspired} & \multirow{2}{*}{Others} \\
\cline{2-16}
 & EN & MI & FMI & PSNR & AG & $Q_{\text{ABF}}$ & SD & SF & $Q_{\text{C}}$ & SCD & CC & SSIM & $Q_{\text{CB}}$ & $Q_{\text{CV}}$ & $Q_{\text{VIFF}}$ & \\ 
\hline
DenseFuse~\cite{li2018densefuse}  & ${\surd}$  &    & ${\surd}$  &    &    & ${\surd}$  &    &    &    & ${\surd}$  &   & ${\surd}$  &    &    &    &  \\ 
FusionGAN~\cite{ma2019fusiongan}  &  ${\surd}$  &    &   &    &    &    &  ${\surd}$  &  ${\surd}$  &    &   & ${\surd}$  & ${\surd}$  &    &    &  ${\surd}$  &  \\ 
U2Fusion~\cite{xu2020u2fusion}  &    &    &   &  ${\surd}$  &    &    &    &    &    & ${\surd}$  & ${\surd}$  & ${\surd}$  &    &    &    &  \\ 
DDcGAN~\cite{ma2020ddcgan}  &  ${\surd}$  &    &   &  ${\surd}$  &  ${\surd}$  &    &  ${\surd}$  &  ${\surd}$  &    &   & ${\surd}$  & ${\surd}$  &    &    &  ${\surd}$  &  \\ 
SDNet~\cite{zhang2021sdnet}  &  ${\surd}$  &    & ${\surd}$  &  ${\surd}$  &  ${\surd}$  &    &    &    &    &   &   &   &    &    &    &   \\ 
RPNNest~\cite{li2021rfn}  &  ${\surd}$  &  ${\surd}$  &   &    &    &  ${\surd}$  &  ${\surd}$  &    &    & ${\surd}$  &   & ${\surd}$  &    &    &  ${\surd}$  &  \\ 
SwinFusion~\cite{ma2022swinfusion}  &    &    & ${\surd}$  &  ${\surd}$  &    &  ${\surd}$  &    &    &    &   &   & ${\surd}$  &    &    &  ${\surd}$  &   \\ 
PIAFusion~\cite{tang2022piafusion}  &    &  ${\surd}$  &   &    &    &  ${\surd}$  &  ${\surd}$  &    &    &   &   & ${\surd}$  &    &    &    &  \\ 
TarDAL~\cite{liu2022target}  &  ${\surd}$  &  ${\surd}$  &   &    &    &    &  ${\surd}$  &    &    &   &   &   &    &    &  ${\surd}$  &  \\ 
LRRNet~\cite{li2023lrrnet}  &  ${\surd}$  &  ${\surd}$  &   &    &    &  ${\surd}$  &  ${\surd}$  &    &    &   &   & ${\surd}$  &    &    &    &   \\ 
DifFusion~\cite{yue2023dif}  &    &  ${\surd}$  &   &    &    &  ${\surd}$  &  ${\surd}$  &  ${\surd}$  &    &   &   & ${\surd}$  &    &    &    &  $\Delta{}E$ \\ 
DIVFusion~\cite{tang2023divfusion}  &  ${\surd}$  &    &   &    &  ${\surd}$  &    &  ${\surd}$  &    &    & ${\surd}$  & ${\surd}$  & ${\surd}$  &    &    &    &   \\ 
DLF~\cite{li2023deep}  &  ${\surd}$  &    &   &    &    &  ${\surd}$  &    &    &    &   &   & ${\surd}$  &  ${\surd}$  &  ${\surd}$  &    &  CE \\ 
CDDFuse~\cite{zhao2023cddfuse}  &  ${\surd}$  &  ${\surd}$  &   &    &    &  ${\surd}$  &  ${\surd}$  &  ${\surd}$  &    & ${\surd}$  &   & ${\surd}$  &    &    &    &  \\ 
MetaFusion~\cite{zhao2023metafusion}  &  ${\surd}$  &  ${\surd}$  &   &    &    &    &    &    &    &   &   & ${\surd}$  &    &    &    &  \\ 
TGFuse~\cite{rao2023tgfuse}  &  ${\surd}$  &  ${\surd}$  & ${\surd}$  &    &    &  ${\surd}$  &  ${\surd}$  &  ${\surd}$  &    &   &   & ${\surd}$  &    &    &    &   \\ 
DDFM~\cite{zhao2023ddfm}  &  ${\surd}$  &  ${\surd}$  & ${\surd}$  &    &    &  ${\surd}$  &  ${\surd}$  &    &    &   &   & ${\surd}$  &    &    &    &  \\ 
SHIP~\cite{zheng2024probing}  &    &  ${\surd}$  &   &    &  ${\surd}$  &  ${\surd}$  &    &  ${\surd}$  &    &   &   & ${\surd}$  &    &    &    &   \\ 
TCMoA~\cite{zhu2024task}  &  ${\surd}$  &    & ${\surd}$  &    &    &    &  ${\surd}$  &    &  ${\surd}$  &   &   & ${\surd}$  &    &  ${\surd}$  &    &  
$Q_{\text{W}}$ \\ 
TextIF~\cite{yi2024text}  &  ${\surd}$  &    &   &    &    &  ${\surd}$  &  ${\surd}$  &    &    & ${\surd}$  &   &   &    &    &    &   \\ 
DDBF~\cite{zhang2024dispel}  &    &  ${\surd}$  &   &    &  ${\surd}$  &    &  ${\surd}$  &    &    &   &   & ${\surd}$  &    &    &    &  \\ 
EMMA~\cite{zhao2024equivariant}  &  ${\surd}$  &    &   &    &  ${\surd}$  &    &  ${\surd}$  &  ${\surd}$  &    & ${\surd}$  &   &  &    &    &    &  \\ 
SeAFusion~\cite{tang2022image}  &  ${\surd}$   &    &   &    &    &  ${\surd}$  &  ${\surd}$   &  ${\surd}$   &    &   &   &    &    &    &    &  \\ 
SuperFusion~\cite{tang2022superfusion}  &    &  ${\surd}$  & ${\surd}$  &    &    &  ${\surd}$  &    &    &    &   &   & ${\surd}$  &    &    &    &  \\ 
PSFusion~\cite{tang2023rethinking}  &  ${\surd}$  &    &   &    & ${\surd}$   &    &  ${\surd}$  &  ${\surd}$  &    & ${\surd}$  &   & ${\surd}$  &    &    & ${\surd}$   &   \\ 
SegMiF~\cite{liu2023multi}  &  ${\surd}$  &    &   &    &    &    &  ${\surd}$  &  ${\surd}$  &    & ${\surd}$  &   &   &    &    &    &   \\ 
PAIF~\cite{liu2023paif}  &    &    &   &    &    &    &    &    &    &   &   &  &    &    &   &  \\ 
TIM~\cite{liu2024task}  &    &  ${\surd}$  & ${\surd}$  &    &    &  ${\surd}$  &    &    &    &   &   &   &    &    &${\surd}$    &   \\ 
SDCFusion~\cite{liu2024semantic}  &  ${\surd}$  &    &   &  ${\surd}$  &    &  ${\surd}$  &  ${\surd}$  &    &    & ${\surd}$  &   &   &    &    & ${\surd}$    &   \\ 
MRFS~\cite{zhang2024mrfs}  &  ${\surd}$  &    &   &    &    &    &  ${\surd}$  &    &    &   &   &   &    &    &  &  \\ 
\hline
Total  &  21  &  13  &  8  &  5  &  7  &  16  &  20  &  10  &  1  &  10  &  4  &  20  &  1  &  2  &  8  &  \\ \hline 
\end{tabular}
}
\label{tab_iqa_metrics}
\end{table*}

\renewcommand\arraystretch{1.66}
\begin{table*}[!t]
\caption{Qualitative comparisons of different VIF methods using conventional evaluation methods on the FMB Dataset. The best and second best results are highlighted in \textbf{bold} and \underline{underline}, respectively.}
\centering
\resizebox{\textwidth}{!}{%
\begin{tabular}{|l|cccc|cccc|cccc|ccc|}
\hline
Method & EN$\uparrow$ & MI$\uparrow$ & FMI$\uparrow$ & PSNR$\uparrow$ & AG$\uparrow$ & $Q^{\text{AB/F}}$$\uparrow$ & SD$\uparrow$ & SF$\uparrow$ & $Q_{C}$$\uparrow$ & SCD$\uparrow$ & CC$\uparrow$ & SSIM$\uparrow$ & $Q_{\text{CB}}$$\uparrow$ & $Q_{\text{CV}}$$\downarrow$ & $Q_{\text{VIFF}}$$\uparrow$ \\ \hline
Visible & 6.522 & 0.569 & \textbf{0.504} & \textbf{60.958} & 4.000 & \textbf{0.722} & 31.886 & 13.820 & \textbf{0.945} & 0.596 & 0.467 & 1.400 & 0.474 & \textbf{1.647} & 0.299 \\
Infrared & 6.874 & 0.569 & \textbf{0.504} & \textbf{60.958} & 1.421 & 0.282 & 44.349 & 4.304 & 0.543 & 0.818 & 0.467 & 1.400 & \textbf{0.836} & 77.137 & 0.140 \\
\hline
DenseFuse & 6.648 & 0.463 & 0.250 & 60.496 & 2.866 & 0.535 & 32.204 & 9.149 & 0.794 & 1.531 & 0.633 & \textbf{1.463} & 0.477 & 16.675 & 0.459 \\
FusionGAN & 6.497 & 0.460 & 0.142 & 60.541 & 2.453 & 0.321 & 28.544 & 8.261 & 0.798 & 1.056 & 0.542 & 1.409 & 0.378 & 23.500 & 0.194 \\
U2Fusion & 6.751 & 0.428 & 0.252 & 60.747 & 3.638 & 0.592 & 32.087 & 10.856 & 0.802 & 1.630 & \textbf{0.649} & \textbf{1.463} & 0.484 & 17.144 & 0.497 \\
DDcGAN & \underline{7.475} & 0.381 & 0.206 & 59.254 & 5.566 & 0.496 & \underline{51.381} & 17.077 & 0.826 & 1.611 & 0.593 & 1.234 & 0.515 & 20.195 & 0.627 \\
SDNet & 6.604 & 0.466 & 0.213 & 60.639 & 4.032 & 0.559 & 34.805 & 13.077 & 0.765 & 1.384 & 0.567 & 1.429 & 0.444 & 14.898 & 0.391 \\
RPNNest & 6.819 & 0.445 & 0.240 & 60.223 & 2.562 & 0.447 & 35.327 & 7.782 & 0.790 & 1.600 & 0.622 & 1.412 & \underline{0.520} & 28.834 & 0.457 \\
SwinFusion & 6.672 & 0.525 & 0.346 & 59.880 & 4.052 & 0.651 & 35.382 & 13.505 & 0.887 & 1.447 & 0.590 & 1.412 & 0.431 & 8.357 & 0.492 \\
PIAFusion & 6.666 & 0.743 & 0.393 & 60.363 & 4.304 & 0.707 & 34.682 & 14.063 & 0.898 & 1.407 & 0.558 & 1.409 & 0.449 & 6.330 & 0.447 \\
TarDAL & 7.017 & 0.467 & 0.177 & 60.222 & 3.565 & 0.443 & 41.279 & 11.643 & 0.774 & 1.626 & 0.603 & 1.393 & 0.488 & 43.449 & 0.431 \\
LRRNet & 6.281 & 0.421 & 0.247 & 60.636 & 3.090 & 0.521 & 26.264 & 10.176 & 0.817 & 1.166 & 0.588 & 1.442 & 0.383 & 14.152 & 0.353 \\
DifFusion & 6.592 & \textbf{0.957} & \underline{0.445} & \underline{60.767} & 4.252 & \underline{0.710} & 33.856 & 14.664 & \underline{0.938} & 0.520 & 0.467 & 1.391 & 0.461 & \underline{2.019} & 0.327 \\
DIVFusion & \textbf{7.566} & 0.388 & 0.181 & 58.835 & 4.922 & 0.471 & \textbf{54.550} & 15.280 & 0.816 & 1.580 & 0.597 & 1.212 & 0.470 & 21.559 & 0.732 \\
DLF & 6.800 & 0.420 & 0.268 & 60.201 & 3.061 & 0.516 & 34.589 & 9.224 & 0.840 & 1.504 & 0.612 & 1.324 & 0.497 & 21.786 & 0.258 \\
CDDFuse & 6.824 & 0.607 & 0.334 & 60.162 & 4.341 & 0.667 & 38.581 & 14.552 & 0.888 & 1.668 & 0.621 & 1.423 & 0.461 & 7.568 & 0.568 \\
MetaFusion & 7.206 & 0.359 & 0.202 & 60.131 & \textbf{6.340} & 0.467 & 42.798 & \textbf{18.545} & 0.774 & 1.660 & 0.622 & 1.276 & 0.461 & 27.289 & \textbf{1.050} \\
TGFuse & 6.610 & 0.442 & 0.371 & 60.302 & 4.239 & 0.674 & 31.724 & 14.146 & 0.879 & 1.135 & 0.527 & 1.387 & 0.388 & 3.233 & 0.471 \\
DDFM & 6.705 & 0.345 & 0.014 & 60.043 & 2.852 & 0.061 & 31.828 & 8.872 & 0.614 & 1.152 & 0.527 & 1.048 & 0.467 & 76.860 & 0.066 \\
SHIP & 6.721 & \underline{0.757} & 0.377 & 60.072 & 4.478 & 0.686 & 35.809 & 14.568 & 0.835 & 1.412 & 0.571 & 1.386 & 0.447 & 7.681 & 0.442 \\
TCMoA & 6.688 & 0.469 & 0.215 & 60.445 & 3.376 & 0.510 & 34.927 & 10.211 & 0.831 & 1.444 & 0.583 & 1.426 & 0.446 & 18.140 & 0.487 \\
TextIF & 6.780 & 0.602 & 0.366 & 60.421 & 4.389 & 0.689 & 36.666 & 14.210 & 0.844 & 1.528 & 0.589 & 1.406 & 0.454 & 4.965 & 0.509 \\
DDBF & 6.428 & 0.500 & 0.269 & 57.502 & 5.435 & 0.576 & 29.938 & 17.942 & 0.835 & 0.958 & 0.493 & 1.208 & 0.286 & 25.104 & 0.303 \\
EMMA & 6.819 & 0.589 & 0.318 & 60.369 & 4.744 & 0.645 & 38.470 & 15.078 & 0.873 & 1.521 & 0.591 & 1.401 & 0.457 & 24.196 & 0.534 \\
\hline
SeAFusion & 6.810 & 0.570 & 0.316 & 59.820 & 4.301 & 0.649 & 37.747 & 13.867 & 0.829 & 1.648 & 0.616 & 1.407 & 0.455 & 11.951 & 0.512 \\
SuperFusion & 6.612 & 0.658 & 0.276 & 60.198 & 3.478 & 0.591 & 33.597 & 11.566 & 0.863 & 1.440 & 0.583 & 1.431 & 0.455 & 11.354 & 0.401 \\
PSFusion & 7.299 & 0.409 & 0.273 & 59.285 & \underline{5.673} & 0.623 & 49.456 & \underline{18.144} & 0.736 & \textbf{1.887} & \underline{0.643} & 1.310 & 0.493 & 20.257 & \underline{0.829} \\
SegMiF & 7.002 & 0.476 & 0.337 & 60.376 & 4.214 & 0.679 & 40.968 & 13.811 & 0.843 & \underline{1.774} & 0.637 & 1.409 & 0.491 & 8.141 & 0.614 \\
PAIF & 6.658 & 0.455 & 0.194 & 60.550 & 2.724 & 0.395 & 35.767 & 8.420 & 0.769 & 1.592 & 0.596 & \underline{1.461} & 0.457 & 23.758 & 0.344 \\
TIM & 6.609 & 0.530 & 0.274 & 60.609 & 3.626 & 0.590 & 30.324 & 12.211 & 0.859 & 1.309 & 0.571 & 1.416 & 0.425 & 11.460 & 0.348 \\
SDCFusion & 6.856 & 0.499 & 0.349 & 59.898 & 4.441 & 0.690 & 37.122 & 14.095 & 0.773 & 1.692 & 0.622 & 1.396 & 0.446 & 7.434 & 0.590 \\
MRFS & 6.859 & 0.508 & 0.302 & 60.107 & 3.666 & 0.611 & 40.490 & 12.407 & 0.854 & 1.291 & 0.554 & 1.414 & 0.455 & 9.455 & 0.489 \\
\hline
\end{tabular}
}
\label{tab_iqa_fmb}
\end{table*}

\renewcommand\arraystretch{1.66}
\begin{table*}[!t]
\caption{Qualitative comparisons of different VIF methods using conventional evaluation methods on the MVSeg Dataset. The best and second best results are highlighted in \textbf{bold} and \underline{underline}, respectively.}
\centering
\resizebox{\textwidth}{!}{%
\begin{tabular}{|l|cccc|cccc|cccc|ccc|}
\hline
Method & EN$\uparrow$ & MI$\uparrow$ & FMI$\uparrow$ & PSNR$\uparrow$ & AG$\uparrow$ & $Q^{\text{AB/F}}$$\uparrow$ & SD$\uparrow$ & SF$\uparrow$ & $Q_{C}$$\uparrow$ & SCD$\uparrow$ & CC$\uparrow$ & SSIM$\uparrow$ & $Q_{\text{CB}}$$\uparrow$ & $Q_{\text{CV}}$$\downarrow$ & $Q_{\text{VIFF}}$$\uparrow$ \\ \hline
Visible & 7.068 & 0.556 & \textbf{0.504} & \textbf{58.430} & 5.301 & \textbf{0.693} & \textbf{60.555} & 15.288 & \textbf{0.930} & 0.930 & 0.446 & 1.274 & \textbf{0.785} & 164.630 & 0.592 \\
Infrared & 5.976 & 0.556 & \textbf{0.504} & \textbf{58.430} & 1.700 & 0.349 & 24.534 & 5.174 & 0.606 & 0.400 & 0.446 & 1.274 & 0.561 & 1346.756 & 0.040 \\
\hline
DenseFuse & 6.912 & 0.483 & 0.238 & 58.100 & 3.863 & 0.560 & 41.499 & 10.496 & 0.813 & 1.590 & 0.564 & \textbf{1.317} & 0.541 & 239.434 & 0.533 \\
FusionGAN & 6.320 & 0.383 & 0.089 & 57.459 & 2.694 & 0.219 & 25.678 & 7.590 & 0.658 & 1.018 & 0.535 & 1.133 & 0.397 & 672.126 & 0.159 \\
U2Fusion & 6.743 & 0.381 & 0.207 & \underline{58.215} & 4.640 & 0.556 & 35.446 & 12.017 & 0.799 & 1.388 & 0.550 & 1.281 & 0.517 & 331.471 & 0.482 \\
DDcGAN & \underline{7.513} & 0.317 & 0.119 & 56.631 & 6.174 & 0.435 & 51.815 & 16.102 & 0.738 & 1.347 & 0.541 & 0.980 & 0.493 & 564.153 & 0.481 \\
SDNet & 6.477 & 0.328 & 0.155 & 57.965 & 5.157 & 0.475 & 27.091 & 14.300 & 0.720 & 1.253 & \underline{0.571} & 1.220 & 0.364 & 565.129 & 0.275 \\
RPNNest & 7.136 & 0.425 & 0.236 & 58.004 & 3.738 & 0.494 & 44.852 & 9.523 & 0.820 & 1.613 & 0.564 & 1.276 & 0.577 & 290.095 & 0.558 \\
SwinFusion & 6.958 & 0.526 & 0.313 & 57.517 & 5.186 & 0.620 & 53.505 & 14.707 & 0.836 & 1.378 & 0.490 & 1.281 & 0.590 & 190.199 & 0.573 \\
PIAFusion & 7.084 & \underline{0.704} & \underline{0.359} & 57.944 & 5.569 & 0.669 & 55.888 & 15.378 & 0.850 & 1.299 & 0.479 & 1.273 & 0.634 & \textbf{144.083} & 0.591 \\
TarDAL & 7.104 & 0.464 & 0.141 & 57.633 & 4.485 & 0.460 & 53.800 & 12.706 & 0.784 & 1.455 & 0.504 & 1.226 & 0.550 & 273.511 & 0.524 \\
LRRNet & 6.978 & 0.439 & 0.174 & 58.043 & 4.149 & 0.486 & 45.054 & 11.662 & 0.774 & 1.355 & 0.509 & 1.266 & 0.566 & 266.412 & 0.489 \\
DifFusion & 7.059 & 0.500 & 0.222 & 57.628 & 5.609 & 0.567 & 52.075 & 14.942 & 0.792 & 1.343 & 0.488 & 1.239 & 0.560 & 202.836 & 0.564 \\
DIVFusion & \textbf{7.518} & 0.391 & 0.103 & 56.773 & 5.118 & 0.415 & 55.009 & 12.955 & 0.744 & 1.572 & 0.544 & 1.064 & 0.571 & 345.761 & 0.628 \\
DLF & 7.062 & 0.420 & 0.236 & 57.924 & 4.165 & 0.500 & 51.289 & 11.012 & 0.826 & 1.388 & 0.492 & 1.177 & 0.605 & 259.714 & 0.460 \\
CDDFuse & 7.153 & 0.597 & 0.282 & 57.814 & 5.589 & 0.636 & 58.903 & 15.797 & 0.813 & 1.514 & 0.499 & 1.274 & 0.620 & 169.613 & 0.656 \\
MetaFusion & 7.167 & 0.361 & 0.154 & 57.627 & \textbf{8.232} & 0.508 & 58.841 & \textbf{21.201} & 0.791 & 1.536 & 0.510 & 0.965 & 0.601 & 269.087 & \underline{0.704} \\
TGFuse & 7.204 & 0.504 & 0.353 & 57.833 & 5.515 & 0.642 & 57.984 & 15.633 & 0.865 & 1.233 & 0.461 & 1.249 & 0.641 & 156.945 & 0.638 \\
DDFM & 7.032 & 0.365 & 0.117 & 57.799 & 4.053 & 0.320 & 43.604 & 10.657 & 0.700 & 1.431 & 0.510 & 1.044 & 0.538 & 565.434 & 0.374 \\
SHIP & 7.046 & \textbf{0.729} & 0.344 & 57.728 & 5.719 & 0.657 & 53.835 & 16.047 & 0.829 & 1.295 & 0.480 & 1.253 & 0.586 & 174.686 & 0.573 \\
TCMoA & 7.021 & 0.399 & 0.168 & 58.083 & 4.181 & 0.473 & 41.687 & 10.910 & 0.778 & 1.421 & 0.535 & 1.283 & 0.519 & 222.025 & 0.493 \\
TextIF & 7.188 & 0.649 & 0.349 & 58.073 & 5.658 & 0.666 & 59.954 & 15.316 & 0.843 & 1.429 & 0.480 & 1.266 & 0.673 & 157.645 & 0.670 \\
DDBF & 6.851 & 0.435 & 0.159 & 55.726 & \underline{6.425} & 0.494 & 42.547 & \underline{18.187} & 0.752 & 1.199 & 0.480 & 1.034 & 0.385 & 373.980 & 0.461 \\
EMMA & 7.168 & 0.539 & 0.274 & 57.788 & 5.892 & 0.606 & \underline{60.367} & 16.265 & 0.832 & 1.336 & 0.479 & 1.251 & 0.625 & 286.076 & 0.658 \\
\hline
SeAFusion & 7.067 & 0.563 & 0.278 & 57.682 & 5.509 & 0.626 & 53.547 & 15.200 & 0.813 & 1.489 & 0.502 & 1.275 & 0.579 & 194.050 & 0.584 \\
SuperFusion & 6.890 & 0.631 & 0.237 & 57.741 & 4.489 & 0.580 & 52.322 & 12.927 & 0.818 & 1.373 & 0.489 & \underline{1.302} & 0.596 & 198.737 & 0.521 \\
PSFusion & 7.335 & 0.451 & 0.274 & 57.858 & 6.274 & 0.654 & 54.354 & 16.665 & 0.812 & \textbf{1.701} & 0.533 & 1.224 & 0.601 & 182.642 & \textbf{0.724} \\
SegMiF & 7.053 & 0.484 & 0.288 & 57.835 & 5.428 & 0.628 & 55.157 & 15.208 & 0.781 & \underline{1.667} & 0.512 & 1.163 & 0.586 & 212.560 & 0.648 \\
PAIF & 6.280 & 0.361 & 0.173 & 57.844 & 3.915 & 0.453 & 30.858 & 10.444 & 0.724 & 1.409 & \textbf{0.589} & 1.235 & 0.416 & 506.658 & 0.316 \\
TIM & 7.099 & 0.613 & 0.256 & 58.210 & 4.840 & 0.597 & 54.607 & 13.596 & \underline{0.892} & 1.398 & 0.486 & 1.276 & \underline{0.689} & 197.483 & 0.562 \\
SDCFusion & 7.174 & 0.540 & 0.329 & 57.785 & 5.708 & \underline{0.670} & 54.512 & 15.140 & 0.818 & 1.501 & 0.497 & 1.260 & 0.581 & \underline{152.521} & 0.671 \\
MRFS & 7.149 & 0.461 & 0.230 & 57.931 & 4.452 & 0.567 & 58.164 & 13.158 & 0.851 & 1.377 & 0.483 & 1.282 & 0.619 & 189.101 & 0.575 \\
\hline
\end{tabular}
}
\label{tab_iqa_mvseg}
\end{table*}

\section{Correlation Analysis}
\label{Correlation Analysis}

Existing VIF papers predominantly utilize conventional evaluation metrics to quantitatively assess their methodologies. According to the taxonomy introduced in prior works~\cite{zhang2023visible, ma2019infrared, liu2011objective}, these metrics can be categorized into four groups. The first group is information theory-based metrics, including Cross-entropy (EN)~\cite{roberts2008assessment}, Entropy (EN)~\cite{roberts2008assessment}, Mutual Information (MI)~\cite{qu2002information}, Feature Mutual Information (FMI)~\cite{haghighat2011non} and Peak Signal to Noise Ratio (PSNR). The second ground is image feature-based metrics, including Average Gradient (AG)~\cite{cui2015detail}, Edge Intensity (EI)~\cite{rajalingam2018hybrid}, $Q_{\text{ABF}}$~\cite{xydeas2000objective}, Standard Deviation (SD)~\cite{ma2019infrared} and Spatial Frequency (SF)~\cite{eskicioglu1995image}. The third group is structural similarity-based metrics, including $Q_{\text{C}}$~\cite{cvejic2005similarity}, $Q_{\text{W}}$~\cite{wang2004image}, Sum of the Correlations of Differences (SCD)~\cite{aslantas2015new}, Correlation Coefficient (CC)~\cite{deshmukh2010image}, Root Mean Square Error (RMSE)~\cite{jagalingam2015review} and Structural Similarity Index (SSIM)~\cite{wang2004image}. The last group is human perception-inspired metrics, including $Q_{\text{CB}}$~\cite{chen2009new}, $Q_{\text{CV}}$~\cite{chen2007human}, $\Delta{}E$~\cite{sharma2005ciede2000}, $Q_{\text{VIFF}}$~\cite{han2013new}. 
Table~\ref{tab_iqa_metrics} presents the conventional evaluation metrics employed in recent VIF papers. Unlike the latest VIF survey~\cite{zhang2023visible}, which selects 13 conventional evaluation metrics without considering their usage in recent VIF papers, our study incorporates 15 conventional metrics that have been utilized in current VIF research. Notably, certain metrics such as EN and RMSE, included in the aforementioned survey paper, were excluded from our study due to their absence in recent VIF literature.

We use the 15 conventional evaluation metrics to evaluate the recent VIF methods on the FMB and MVSeg datasets, and Table~\ref{tab_iqa_fmb} and Table~\ref{tab_iqa_mvseg} show the experimental results. It can be observed that almost all the latest methods (their papers are published in 2024), such as TCMoA, TextIF, DDBF, EMMA, TIM, SDCFusion and MRFS, did not show advances compared with previous VIF methods. Besides, we observe that the SSIM metric that is widely used in the recent VIF papers (20 out of 30) show that the best performing method is the DenseFuse on both FMB and MVSeg datasets, and this method is proposed by the paper even published in 2019. The reason is that in the publications of these methods, different testing images and different evaluation metrics are selected for evaluation. Such phenomenon has also been pointed out in the latest VIF survey~\cite{zhang2023visible}, and is more severe in VIF methods published in the past year.

\begin{figure*}[!h]
    \centering
        \includegraphics[width=\linewidth]{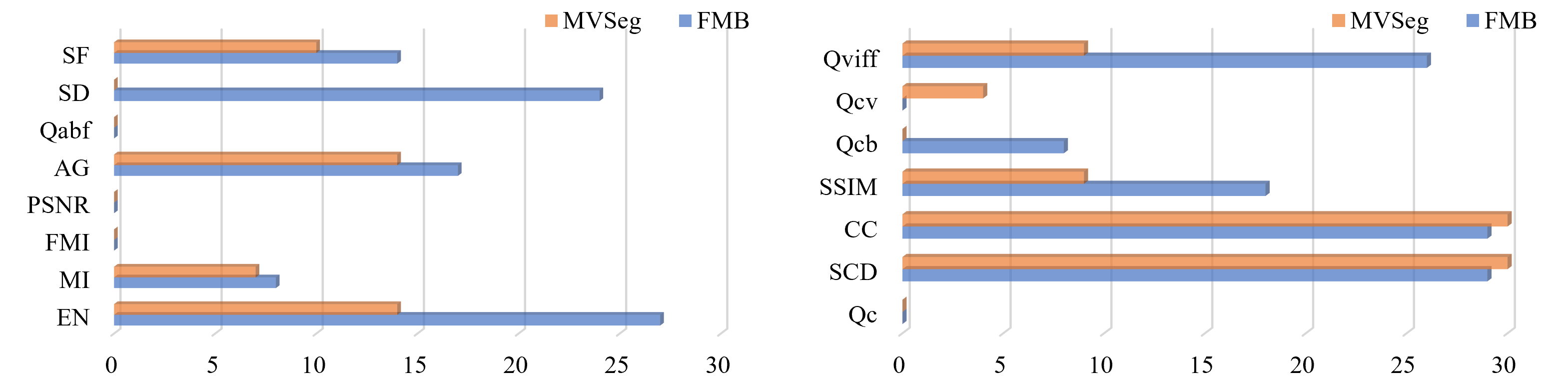}
        \vspace{-10pt}
        \caption{
        The number of VIF methods capable of enhancing the Visible images on the FMB and MVSeg datasets. Notably, none of the evaluated VIF methods succeeded in improving the Visible images across 5 and 6 evaluation metrics, respectively, on the FMB and MVSeg datasets.
        }
    \label{fig_iqa}
\end{figure*}

\renewcommand\arraystretch{1.5}
\begin{table*}[!t]
\caption{Correlation analysis between the SEA and existing evaluation metrics. The best and second best results are highlighted in \textbf{bold} and \underline{underline}, respectively.}
\centering
\resizebox{\textwidth}{!}{%
\begin{tabular}{|l|cccc|cccc|cccc|ccc|}
\hline
Dataset & EN & MI & FMI & PSNR & AG & $Q_{\text{ABF}}$ & SD & SF & $Q_{\text{C}}$ & SCD & CC & SSIM & $Q_{\text{CB}}$ & $Q_{\text{CV}}$ & $Q_{\text{VIFF}}$ \\ 
\hline
FMB &0.163 &0.342 &0.376 &0.122 &0.313 &\textbf{0.503} &0.177 &-0.150 &0.299 &0.359 &0.303 &0.269 &0.040 &-0.074 &\underline{0.382} \\
MVSeg &0.139 &0.346 &0.299 &0.305 &0.097 &\underline{0.357} &0.303 &0.311 &0.251 &0.061 &-0.136 &0.236 &0.355 &0.265 &\textbf{0.386} \\ \hline
Mean & 0.151 & 0.344 & 0.338 & 0.214 & 0.205 & \textbf{0.430} & 0.240 & 0.081 & 0.275 & 0.210 & 0.084 & 0.252 & 0.198 & 0.096 & \underline{0.384} \\ \hline
\end{tabular}
}
\label{tab_corr}
\end{table*}

It is noteworthy that all compared VIF methods exhibit inferior performance to using the Visible images across various evaluation metrics, particularly on the MVSeg dataset. As depicted in Figure~\ref{fig_iqa}, the number of VIF methods surpassing the Visible on the FMB and MVSeg datasets is limited. In general, VIF methods yield better results on the FMB dataset compared to the MVSeg dataset. Specifically, on the FMB dataset, every evaluated VIF method underperforms relative to the Visible when assessed with FMI, PSNR, $Q_{\text{ABF}}$, $Q_{\text{C}}$, and $Q_{\text{CV}}$. Similarly, on the MVSeg dataset, all VIF methods fall short of the Visible's performance according to FMI, PSNR, $Q_{\text{ABF}}$, SD, $Q_{\text{C}}$, and $Q_{\text{CB}}$, accounting for 40\% of the 15 conventional evaluation metrics. These findings suggest a relationship between conventional evaluation metrics and our proposed SEA, underscoring that no method evaluated demonstrates an improvement over the Visible on the MVSeg dataset.

To gain a comprehensive understanding of the relationship between conventional evaluation metrics and our proposed SEA, we utilize statistical correlation measures to examine their consistency. Specifically, we employ Kendall's $\tau$ rank correlation coefficient~\cite{kendall1938new} to measure the similarity between fusion metrics. As indicated in Table~\ref{tab_corr}, the metrics most strongly correlated with our SEA are $Q_{\text{ABF}}$~\cite{xydeas2000objective} and $Q_{\text{VIFF}}$~\cite{han2013new}. The $Q_{\text{ABF}}$ metric evaluates the preservation and integration of edge information from source images into the final fused image, while $Q_{\text{VIFF}}$ assesses visual information fidelity, aligning closely with human visual perception capabilities. This correlation analysis is supported by qualitative results, as illustrated in Figures~\ref{fig_motivation}~and~\ref{fig_sub_assess}. Poor visual quality in images impacts not only edge information and visual information fidelity but also semantic information. Therefore, considering $Q_{\text{ABF}}$ and $Q_{\text{VIFF}}$ when segmentation labels are unavailable broadens the applicability of our evaluations.

\section{Conclusion}
\label{Conclusion}

This paper presents a Segmentation-oriented Evaluation Approach (SEA) for assessing Visible and Infrared Image Fusion (VIF) methods using universal segmentation models. The SEA addresses the critical challenge of evaluating VIF methods in the absence of ground-truth fused images, offering a robust and universally applicable solution across diverse VIF datasets.

Experimental results highlight the SEA's ability to distinguish between high-quality and low-quality fusion methods, revealing that only a few recent VIF methods achieve significant performance gains. The correlation analysis further supports the validity of SEA by showing strong alignment with conventional metrics, highlighting its reliability as an evaluation tool.

The contributions of this work are threefold. First, it introduces a novel and practical method for evaluating VIF methods, overcoming the traditional limitations of ground-truth unavailability. Second, the SEA is universally applicable, making it adaptable to various VIF datasets and tasks, thus providing a more holistic evaluation framework. Third, the comparative study offers a comprehensive analysis of recent VIF methods, setting a new benchmark for future research in this domain.

Future research could explore two main directions. First, integrating SEA with emerging vision-language models to leverage the rich semantic information available in textual descriptions, potentially leading to more accurate fusion evaluations. Second, developing new state-of-the-art VIF models that excel under the proposed SEA evaluation metric, pushing the boundaries of current VIF performance.

\ifCLASSOPTIONcaptionsoff
  \newpage
\fi

{\small
\bibliographystyle{IEEEtran}
\bibliography{reference}
}

\end{document}